\documentclass{article} 
\usepackage[final]{colm2026/colm2026_conference}

\usepackage{microtype}
\usepackage{hyperref}
\usepackage{url}
\usepackage{booktabs}

\usepackage{lineno}

\definecolor{darkblue}{rgb}{0, 0, 0.5}
\hypersetup{colorlinks=true, citecolor=darkblue, linkcolor=darkblue, urlcolor=darkblue}

\usepackage{makecell}
\usepackage{amsfonts}
\usepackage{amssymb}
\usepackage{amsmath}
\usepackage{alltt}
\usepackage{pifont}
\usepackage{subcaption}
\usepackage{multirow}
\usepackage{tikz}
\usepackage{pgfplots}
\usepackage{tabularx}
\usepackage{listings}
\usepackage{enumitem}

\pgfplotsset{compat=1.18}
\usetikzlibrary{shapes, positioning}

\usepackage{xcolor}
\definecolor{red}{rgb}{0.957,0.498,0.447}
\definecolor{lightred}{rgb}{0.965,0.702,0.675}
\definecolor{green}{rgb}{0.553,0.824,0.773}
\definecolor{lightgreen}{rgb}{0.729,0.890,0.863}
\definecolor{blue}{rgb}{0.498,0.698,0.835}
\definecolor{lightblue}{rgb}{0.702,0.820,0.906}
\definecolor{orange}{rgb}{0.969,0.718,0.427}
\definecolor{lightorange}{rgb}{0.988,0.831,0.631}
\definecolor{purple}{rgb}{0.749,0.737,0.855}
\definecolor{keywordcolor}{rgb}{0.7, 0.1, 0.1}   
\definecolor{tacticcolor}{rgb}{0.0, 0.1, 0.6}    
\definecolor{commentcolor}{rgb}{0.3, 0.5, 0.3}   
\definecolor{symbolcolor}{rgb}{0.0, 0.1, 0.6}    
\definecolor{sortcolor}{rgb}{0.1, 0.5, 0.1}      
\definecolor{rulecolor}{rgb}{0, 0, 0}
\definecolor{attributecolor}{rgb}{0.7, 0.1, 0.1} 

\usepackage[linesnumbered,ruled,vlined]{algorithm2e}
\SetKwInOut{KwIn}{Input}
\SetKwInOut{KwOut}{Output}

\SetCommentSty{mycommfont}

\title{MechMath: Sorrifier-Driven Formal Decomposition Workflow for Automated Theorem Proving}

\author{Ruichen Qiu\thanks{Equal contribution.} \\
SKLMS, Academy of Mathematics and Systems Science, CAS, Beijing, 100190, China\\
School of Advanced Interdisciplinary Sciences, UCAS, Beijing, 101408, China \\
\texttt{qiuruichen20@mails.ucas.ac.cn} \\
\AND
Yichuan Cao$^*$, Junqi Liu$^*$, Dakai Guo, Xiao-Shan Gao, Lihong Zhi \& Ruyong Feng\thanks{Corresponding author.} \\
SKLMS, Academy of Mathematics and Systems Science, CAS, Beijing, 100190, China \\
School of Mathematical Science, UCAS, Beijing, 101408, China\\
\texttt{ryfeng@amss.ac.cn}
}

\begin{document}

\def\lstlanguagefiles{lstlean.tex}
\lstdefinestyle{lean}{
    escapeinside=``,
    language=lean,
    basicstyle=\scriptsize\ttfamily
}

\newcommand{\orangehl}[2][lightorange!50]{%
    \setlength{\fboxsep}{0.5pt}%
    \raisebox{0pt}[0pt][0pt]{\colorbox{#1}{\strut #2}}%
}
\newcommand{\bluehl}[2][lightblue!50]{%
    \setlength{\fboxsep}{0.5pt}%
    \raisebox{0pt}[0pt][0pt]{\colorbox{#1}{\strut #2}}%
}
\newcommand{\redhl}[2][lightred!70]{%
    \setlength{\fboxsep}{0.5pt}%
    \raisebox{0pt}[0pt][0pt]{\colorbox{#1}{\strut #2}}%
}
\newcommand{\greenhl}[2][lightgreen!70]{%
    \setlength{\fboxsep}{0.5pt}%
    \raisebox{0pt}[0pt][0pt]{\colorbox{#1}{\strut #2}}%
}

\ifcolmsubmission
\linenumbers
\fi

\maketitle

\begin{abstract}
Recent advances in large language models (LLMs) and LLM-based agents have substantially improved the capabilities of automated theorem proving.
However, for problems that require complex mathematical reasoning, current systems seldom succeed in their initial attempt, necessitating iterative adjustments to their proof strategies.
Existing approaches for handling failed attempts typically either iteratively fix errors within the proof or discard the entire proof and regenerate it from scratch.
The former leads to progressively longer contexts, which degrade the model's ability to attend to the remaining unresolved subproblems, while the latter is inefficient, as it may abandon mostly correct reasoning due to localized errors.
To address this dilemma, we present \textit{MechMath}, an agent system centered on a Sorrifier-driven formal decomposition paradigm.
By leveraging the \texttt{sorry} placeholder in Lean to precisely isolate unresolved subgoals while preserving the surrounding verified proof structure, MechMath extracts each failed subproblem into a clean, self-contained context and resolves it independently. 
This avoids both the waste of full regeneration and the excessive context length induced by repeated repairs.
Experimental results on challenging mathematical competition benchmarks, including IMO 2025, Putnam 2025, miniF2F, and a subset of ProverBench, demonstrate that our agent achieves significant advantages in proving efficiency.
\end{abstract}

\section{Introduction}

Mathematical reasoning has long been one of the core research objectives in artificial intelligence. In recent years, with the rapid advancement of large language models (LLMs), their ability to handle complex mathematical reasoning tasks at the natural language level has improved significantly. State-of-the-art systems have achieved gold-medal performance in high school mathematics competitions (e.g., IMO 2025), while systems such as DeepMind's Aletheia~\citep{feng2026towards} are progressively approaching the frontier of tackling research-level problems.

Despite these advances, theorem proving in natural languages is still plagued by verification difficulties, leading to potential logical hallucinations and erroneous conclusions, and it lacks scalable automated validation~\citep{chen2025seedprover, chen2025seedprover15}. 
In contrast, formal languages such as Lean, Rocq, and Isabelle offer superior logical rigor and verifiability.
In formal math, previous neural theorem proving research primarily focused on training a single model for theorem proving tasks. 
Although such approaches achieved remarkable progress~\citep{lin2025goedel2}, they still show limitations in long-horizon reasoning tasks (e.g., the failure in Putnam 2025 in Table~\ref{tab:putnam}).
As foundation models continue to improve and agent frameworks gain increasing attention, research focus is gradually shifting toward agent systems for formal theorem proving that integrate strategic planning, tool use, and formal verification.

In this emerging agent paradigm, formal proof systems \citep{chen2025seedprover15, varambally2026hilbert} rely on iterative strategies to resolve complex problems. 
The most direct approach involves error correction, where the agent continuously modifies the formal proof script.
However, this frequently leads to contextual explosion, which severely compromises the model's attentional focus on unresolved errors. 
To mitigate this, subsequent research has shifted towards decomposition-based strategies. 
Upon failure, they discard the incorrect proof and regenerate an informal sketch to partition the problem into multiple sub-tasks. 

\input{app/comparison}

However, this informal decomposition paradigm introduces a significant inefficiency.
Whenever a proof attempt fails, the entire formal proof is discarded, and the agent restarts by re-analyzing the problem in natural language and generating new sub-problems (left part of Figure~\ref{fig:comparison}).
On the one hand, the model can often generate an informal sketch whose high-level reasoning structure is largely sound, indicating a potential capacity to generate insightful proofs.
On the other hand, Lean enforces strict logical and syntactic correctness, so even minor omissions or missteps can cause proof compilation to fail.
In such cases, discarding a nearly complete proof and restarting the decomposition is computationally wasteful, both in time and in resources.

This limitation motivates our formal decomposition approach, which preserves formal proof structure and avoids redundant re-analysis.
Using \texttt{sorry}~\footnote{In this paper, words in typewriter font are Lean commands.} placeholder, our approach isolates and resolves localized errors without discarding surrounding correct code (right part of Figure~\ref{fig:comparison}).
Based on this principle, we propose a novel agent system, \textbf{MechMath}, which adopts an informal-to-formal proving pipeline and employs a formal decomposition module, \textbf{Sorrifier}, as illustrated in Figure~\ref{fig:frame}.
Experimental results on challenging mathematical competition benchmarks, including IMO 2025, Putnam 2025, miniF2F, and a subset of ProverBench, demonstrate that our agent achieves significant gains in proving efficiency.

\section{Related Works}
\label{sec:related}

\textbf{Formal Mathematics.} 
Formal mathematics is the representation of mathematical definitions, theorems, and proofs in a rigorous formal language that can be processed by a computer~\citep{harrison1996formalized}.
The key benefit is that correctness is guaranteed, avoiding the risk of human error.
A diverse ecosystem of formal languages and proof assistants has been developed to support this endeavor, such as Rocq and Isabelle~\citep{kaliszyk2020survey}.
Among these, Lean~\citep{moura2021lean}
has emerged as one of the most actively developed and community-driven systems.
The hallmark of the Lean ecosystem is Mathlib~\citep{mathlib}, a unified community-maintained library of formalized mathematics, which contains a vast and growing collection of definitions and theorems.
Refer to Appendix~\ref{app:lean} for more details.

%
%
\textbf{Neural Theorem Proving.}
Previous research on neural theorem proving has largely focused on training a single model, which can be broadly categorized into two main paradigms: step-wise interaction~\citep{xin2025bfs, xin2024deepseek, xin2025deepseek15, hubert2025olympiad} and whole-proof generation~\citep{wang2025kimina, ren2025deepseek2, lin2025goedel, lin2025goedel2}.
Acknowledging the constraints of single-model approaches in long-horizon reasoning and complex proof structures, recent research has shifted toward agent-based paradigms that augment foundation models with tool-use capabilities for the autonomous resolution of complex tasks~\citep{luo2025large}.
In the domain of formal mathematics, a growing body of work explores agentic workflows that augment prover models with additional components to guide proof construction, such as informal reasoning modules~\citep{varambally2026hilbert} and interactive Lean toolchains~\citep{dressler2025lean,breen2025axprover}.
Beyond these enhancements, more advanced paradigms have been proposed, including the training of formal provers through large-scale agentic reinforcement learning~\citep{chen2025seedprover15}, adopting autonomous multi-agent ensemble architectures~\citep{axiom2025from}, and repurposing general coding agents as formal reasoners~\citep{liu2026numina}.


\section{Methods}
\label{sec:methods}
\input{app/frame}
This section presents the architecture and workflow of \textbf{MechMath}. 
To ensure structural clarity, we first introduce core components of our system: LLMs, Lean toolkit, and Sorrifier.
Then, we detail the iterative proof synthesis algorithm that orchestrates these components.

\subsection{Components}
\textbf{LLMs.}
The agent employs three specialized LLMs, each of which has a different functional role. 
The \textbf{Reasoner} is an LLM with a strong mathematical reasoning ability.
Its responsibilities include:
(1) generating informal proof sketches for target statements; and
(2) during subgoal extraction, synthesizing 
necessary premises from the local goal at a given \texttt{sorry} location
to derive the corresponding subgoal.
The \textbf{Verifier} is a general-purpose LLM, tasked with verification and evaluation. Its functions encompass:
(1) evaluating the quality of informal proofs; and
(2) verifying the accuracy and logical soundness of the extracted subgoals.
The \textbf{Prover} is specialized in completing Lean proofs. It translates informal proof sketches into formal Lean proofs and revises incorrect proofs based on feedback.

\textbf{Lean Toolkit.}
Beyond LLMs, we integrate the essential Lean-based utilities required by the agent, including 
(1) the \textbf{Lean compiler}, which checks the correctness of a Lean proof, returns detailed diagnostic feedback, and proof states; and 
(2) the \textbf{Lean data search tools}, which retrieve relevant theorems from Mathlib to assist in constructing proofs.

\textbf{Sorrifier.}
\label{sec:sorrifier}
In Lean, the \texttt{sorry} placeholder allows the compiler to accept the validity of a local lemma without requiring its proof to be fully constructed. 
Replacing erroneous components of a Lean proof with \texttt{sorry}, which is a process referred to as \textit{sorrifying} a faulty proof~\citep{ospanov2025apollo}, can transform an invalid proof into a syntactically valid yet incomplete one. 
This enables the user to focus exclusively on completing the missing parts.

Motivated by this idea, we present the \textbf{Sorrifier}, the central innovation of our pipeline.
This tool repeatedly substitutes incorrect proof blocks with \texttt{sorry} placeholders, continuing this process until the full proof is successfully compiled in Lean.
Building upon this principle, we move beyond the direct approach of discarding an entire failed Lean proof. 
Instead, we isolate and remove the erroneous portions of the proof and substitute them with \texttt{sorry} tactics, thereby producing a formally accepted but incomplete proof. 

To fully realize this goal of maximum retention, the Sorrifier's algorithm must navigate the challenge of cascading errors. 
In dependent type systems like Lean, a single logical flaw often corrupts the local context, triggering a massive cascade of errors in subsequent, otherwise correct code. 
If all reported errors were processed simultaneously, it would lead to unnecessary deletion of valid proof segments. 
Therefore, the Sorrifier is designed to operate through a highly targeted, iterative pipeline.

Specifically, upon encountering compilation errors, the Sorrifier surgically targets the \textit{innermost} reported error. 
If the error lies within a \texttt{have} block, it replaces only the innermost failing proof with \texttt{sorry}. 
If the error occurs elsewhere, the proof is truncated exactly at the erroneous line, and \texttt{sorry} is appended to close the state. 
Crucially, after a single modification is made, the Lean environment is immediately recompiled, instantly removing any downstream cascading errors tied to the previously broken context.
This step-by-step process repeats until the code
is compiled successfully. 
The final output is a structurally valid Lean proof containing a few \texttt{sorry} placeholders. 
Consequently, subsequent efforts can focus solely on resolving these localized subgoals, significantly reducing the overall complexity of the remaining theorem-proving tasks. A detailed algorithm is provided in the Appendix~\ref{app:sorrifier}.

\subsection{Algorithm}
By imitating the way human experts write Lean proofs, the algorithm can be broadly divided into four steps, as illustrated in Figure \ref{fig:frame}.
First, an informal solution is generated and then iteratively refined through a feedback loop in which the Reasoner produces and revises the solution, while the Verifier evaluates it. 
Based on the final informal solution, the Prover generates a formal proof and revises it in response to feedback from Lean.
If the proof cannot be passed
after several revisions, the Sorrifier processes the formal proof to extract the unresolved subgoals.
Each subgoal is handled using the same three-step procedure, and this process repeats until all subgoals have been successfully fulfilled.

\subsubsection{Informal Proof}
When constructing Lean proofs, human experts  typically begin with  an informal proof sketch in natural language before translating  it into a formal proof, as natural language is often better suited to exploratory reasoning.
Inspired by this observation, our agent follows  a similar strategy: it first generates an informal reasoning sketch, which then guides the construction of the  formal proof.
To obtain a coherent and logically structured informal solution, we employ a generate-and-verify loop  between the Reasoner and Verifier, following the approach of \citet{huang2025winning}. 
The Reasoner first generates an informal solution, and  the Verifier evaluates its strategic soundness and logical correctness. 
Based on this assessment, the Reasoner refines the solution, after which a new evaluation begins. 
This iterative refinement process continues until the Verifier accepts the solution as correct or a predefined maximum number of iterations has been reached. 
To improve the reliability of verification, each candidate solution is independently evaluated three times by the Verifier and is accepted only if all three evaluations judge it correct.



\subsubsection{Formal Proof}
Based on the informal proof generated in the previous step, the Prover attempts to write the corresponding Lean proof step-by-step. 
For complex problems, it is challenging to produce a fully correct proof in a single attempt. 
In such cases, error messages returned by the Lean compiler are collected and provided to the Prover to fix errors. 
The error fixing process proceeds over multiple rounds. 
Although each round typically resolves some errors, for longer proofs, the Lean compiler may generate hundreds of error messages in a single pass, making it difficult for the prover to address all issues simultaneously. 
Moreover, fixing an error may introduce new ones. 
Therefore, it is unrealistic to expect that a continuously running fix–verify cycle will eventually correct all errors in the proof. 
After a predefined number of rounds, the proof is passed to the subgoals splitting part. 

\subsubsection{Subgoal Split}
\label{main:subgoal_split}
For the proof processed by the Sorrifier, each place filled by \texttt{sorry} will be converted into a new subgoal, and the aforementioned process will be repeated for the subgoal. 
The specific process is divided into the following three steps, which are illustrated in Figure~\ref{fig:subgoal}.

\input{app/subgoal}
\textbf{Sorrify.}
Given a faulty proof that remains unresolved after several rounds of error correction, the Sorrifier iteratively isolates and replaces erroneous components with \texttt{sorry}.
 This is achieved by surgically targeting the innermost reported error and repeating this process until a successful compilation is achieved. 
The resulting output is a structurally valid Lean proof that passes the compiler but remains incomplete, containing only a limited number of \texttt{sorry} placeholders.
Consider, for example, the proof shown in Figure~\ref{fig:subgoal}. 
A type mismatch error occurs in the \texttt{have} \texttt{step2} of the formal proof. 
In response, the Sorrifier replaces this block with \texttt{sorry}, thus producing a sorried proof.


\textbf{Extract.}
Building upon this sorried proof, our pipeline further facilitates modular reasoning by extracting localized subgoals from each \texttt{sorry} placeholder. 
At each occurrence of a \texttt{sorry} within the Lean proof script, the corresponding tactic state and proof goal are extracted, as illustrated in the orange block of Figure~\ref{fig:subgoal}. 
The Reasoner is then tasked with identifying which hypotheses and contextual conditions in the current tactic state are relevant to the proof goal. 
Irrelevant hypotheses are discarded, and only those that are essential are retained.
Based on this analysis, the model formulates a self-contained subgoal. 
In this manner, complex proofs can be decomposed into smaller, independently verifiable components, thereby streamlining subsequent theorem-proving efforts.



To ensure the reliability and effectiveness of the subgoal extraction process, the subgoals extracted by the Reasoner must undergo a two-stage evaluation by the Verifier before being integrated into the proof pipeline.
The first stage involves syntactic verification using Lean, which checks whether the extracted subgoals are well-formed and adhere to the syntactic conventions of the formal language, ensuring that the subgoals can be processed in subsequent steps without compilation errors.

Following syntactic validation, the extracted subgoals are passed to the Verifier for semantic evaluation. 
This stage assesses the extraction correctness and the logical soundness of the subgoals. 
Extraction correctness evaluates whether they accurately reflect the intended reasoning steps.
Subgoals that are incorrectly extracted deviate from the original proof context and are likely to fail to reassemble into the complete proof. 
Logical soundness measures whether the subgoal can be proven, usually checked by whether there are obvious counterexamples or essential hypotheses missing.
Subgoals that contain logical errors may result from a flawed proof approach, as illustrated in the Appendix~\ref{app:logic_error}. 
They will disrupt the subsequent proof loop, potentially leading to infinite cycles or failed proof attempts. 

\textbf{Assemble.}
Once a subgoal is successfully proven, it is reassembled into the final proof structure, thereby contributing to the completion of the original theorem. 
As illustrated in Step 3 of Figure~\ref{fig:subgoal}, this reintegration is achieved by replacing each \texttt{sorry} placeholder with a corresponding \texttt{apply} tactic. It allows the Lean compiler to index the relevant subgoal at each proof position, enabling the proof to be constructed incrementally. 
In this way, the mechanism ensures that each verified subgoal is incorporated at the correct position, maintaining a modular and incremental proof development process that complements the extraction mechanism described above.
Refer to Appendix~\ref{app:eg_proof} for a real example. 

\subsubsection{Subgoal Process}
\label{sec:subgoal}
Having established the mechanisms for extracting and reintegrating subgoals, we now describe how each subgoal is processed. 
These subgoals undergo the aforementioned proof pipeline recursively. 
If a subgoal cannot be proven directly, it is passed through the Sorrifier, which generates further nested subgoals. 
This iterative decomposition continues until all subgoals are resolved, thereby ensuring that the original theorem is ultimately proven through a structured, hierarchical refinement process.
Since any faulty proof can be sorrified and converted into subgoals, the pipeline may otherwise iterate indefinitely. 
To prevent infinite recursion, we introduce a termination mechanism. 
When a subgoal yields a nested subgoal identical to itself, we deem the proof to have made no progress and revert to a formal proving attempt. 
If the number of such retries reaches a predefined threshold, the subgoal is considered unprovable, and the overall proving process terminates.


\section{Experiments}


\begin{table}[t]
    \scriptsize
    \centering
    \begin{subtable}{\linewidth}
    \subcaption{Results of single models (Pass@32)}
    \begin{tabular*}{\linewidth}
    {@{\hspace{5pt}\extracolsep{\fill}} l c c c c c c c c c c c c @{\hspace{5pt}}}
        \toprule
         & A1 & A2 & A3 & A4 & A5 & A6 & B1 & B2 & B3 & B4 & B5 & B6\\
        \toprule
        Goedel Prover V2 32B & \ding{55} & \ding{55} & \ding{55} & \ding{55} & \ding{55} & \ding{55} & \ding{55} & \ding{55} & \ding{55} & \ding{55} & \ding{55} & \ding{55} \\
        Gemini 3.1 Pro & \ding{55} & \ding{55} & \ding{55} & \ding{55} & \ding{55} & \ding{55} & \ding{55} & \ding{55} & \ding{55} & \ding{55} & \ding{55} & \ding{55} \\
        Claude 4.5 Opus & \ding{55} & \ding{55} & \ding{55} & \ding{55} & \ding{55} & \ding{55} & \ding{55} & \ding{55} & \ding{55} & \ding{55} & \ding{55} & \ding{55} \\
        \bottomrule
    \end{tabular*}
    \end{subtable}
    ~\\
    \begin{subtable}{\linewidth}
    \subcaption{Quantitative results of agents}
    \begin{tabular*}{\linewidth}
    {@{\hspace{5pt}\extracolsep{\fill}} l c c c c c c c c c c c c c @{\hspace{5pt}}}
        \toprule
         & A1 & A2 & A3 & A4 & A5 & A6 & B1 & B2 & B3 & B4 & B5 & B6 & Avg\\
        \toprule
        \multicolumn{14}{c}{Time Expenditure (min)} \\
        \midrule
        Hilbert & - & - & - & - & - & - & - & - & 228 & - & - & - & 228\\
        Seed & 60 & 30 & 120 & 240 & - & 240 & 540 & 360 & 30 & 120 & 240 & 180 & 196\\
        Axiom & 110 & 180 & 165 & 107 & 518 & 259 & 270 & 65 & 43 & 112 & 254 & 494 & 187\\
        Numina& 27 & 81 & 30 & 169 & 2040 & 89 & 55 & 136 & 30 & 308 & 88 & 797 & 165\\
        \cmidrule(lr){1-14}
        \textbf{MechMath-Easy} & \textbf{18} & 45 & 106 & 114 & - & \textbf{79} & - & \textbf{41} & 57 & \textbf{41} & 104 & - & {67.2} \\
        \textbf{MechMath-Hard} & 21 & 179 & 108 & \textbf{64} & - & 82 & 132 & \textbf{41} & 148 & 126 & 258 & \textbf{93} & \textbf{114}\\
        \toprule
        \multicolumn{14}{c}{Cost (\$)}\\
        \midrule
        Hilbert & - & - & - & - & - & - & - & - & 37.9 & - & - & - & 37.9\\
        Numina & 50 & 50 & 50 & 50 & 1000 & 50 & 50 & 50 & 50 & 50 & 50 & 300 & 72.7\\
        \cmidrule(lr){1-14}
        \textbf{MechMath-Easy} & 1.53 & 4.89 & 11.6 & 19.9 & - & 5.48 & - & 5.78 & 7.40 & 7.69 & 25.6 & - & {9.99} \\
        \textbf{MechMath-Hard} & 4.21 & 29.0 & 19.5 & 10.5 & - & 4.91 & 29.8 & 3.72 & 15.4 & 20.1 & 57.0 & 9.01 & \textbf{18.5}\\
        \toprule
        \multicolumn{14}{c}{Proof Tree Complexity (\#Lemmas)}\\
        \midrule
        Hilbert & - & - & - & - & - & - & - & - & 48 & - & - & - & 48\\
        Seed & 27 & 18 & 23 & 28 & - & 38 & 21 & 48 & 26 & 18 & 103 & 185 & 49\\
        Axiom & 23 & 26 & 78 & 32 & 52 & 28 & 49 & 28 & 11 & 23 & 66 & 30 & 36\\
        Numina & 1 & 5 & 21 & 11 & 77 & 37 & 6 & 14 & 9 & 10 & 33 & 41 & 17\\
        \cmidrule(lr){1-14}
        \textbf{MechMath-Easy} & \textbf{1} & \textbf{4} & \textbf{7} & 15 & - & 4 & - & 8 & \textbf{8} & 11 & 39 & - & {11} \\
        \textbf{MechMath-Hard} & \textbf{1} & 20 & 9 & \textbf{10} & - & \textbf{1} & 30 & \textbf{2} & 9 & 17 & 45 & \textbf{10} &\textbf{14}\\
        \bottomrule
    \end{tabular*}
    \end{subtable}
    \caption{Putnam 2025 benchmark. (a) A single prover model or a general LLM cannot solve any problem within 32 attempts. (b) MechMath achieves high efficiency on 11 solved problems: shorter time expenditure (6/11), lower cost (10/11), and fewer theorem counts (8/11). The averages of Axiom and Numina are computed exclusive of A5.}
    \label{tab:putnam}
\end{table}

To comprehensively evaluate \textbf{MechMath}, we conduct experiments on 12 problems from the 2025 Putnam Competition, 4 from the 2025 IMO, the widely-used miniF2F~\citep{zheng2022minif2f} test set, and a subset of ProverBench~\citep{ren2025deepseek2}.
In addition, we perform ablation studies to assess the necessity of each component and the compatibility of different underlying models.

\textbf{Experiment settings}.
We employ Gemini 3.1 Pro~\citep{google2026gemini31pro} as the unified reasoner, verifier, and prover. 
The knowledge cutoff point for this model is January 2025 \footnote{Report on \url{https://ai.google.dev/gemini-api/docs/models/gemini-3.1-pro-preview}.}, ensuring that it has no exposure to Putnam 2025 and IMO 2025. 
We employ the Kimina Lean server~\citep{santos2025kiminaleanserver} for the verification of the Lean code and use LeanDex and Loogle~\footnote{LeanDex: \url{https://leandex.projectnumina.ai/}, Loogle: \url{https://loogle.lean-lang.org/}} for the retrieval of the theorem. 
All experiments are conducted under Lean v4.15.0 (only for miniF2F) and v4.26.0 (for others).
We compare our method against several state-of-the-art baselines:
(1) single models within 32 attempts, including Goedel Prover V2 32B~\citep{lin2025goedel2}, Gemini 3.1 Pro, Claude 4.5 Opus~\citep{claude2025claude45};
(2) formal math agents, including Hilbert~\citep{varambally2026hilbert}, Aristotle~\citep{achim2025aristotle}, Axiom~\citep{axiom2025from}, Seed-Prover 1.5 (Seed)~\citep{chen2025seedprover15}, and Numina-Lean-Agent (Numina)~\citep{liu2026numina}. 
For the open-source Hilbert baseline, we adopt Gemini-3.1-Pro-Preview as the reasoning model to ensure a fair experimental setup. 
For the remaining baselines, we report the performance as documented in their original documents.
More details are given in Appendix~\ref{app-exp}.
The code for reproduction is at \url{https://github.com/MechMath/MechMath-v1}.

\textbf{Easy/Hard mode.}
We categorize our framework into ``easy'' and ``hard'' modes.
The former is the basic pipeline we described in Section~\ref{sec:methods}.
The latter incorporates two specialized modules to handle Lean-specific failures. First, a retriever-based tool addresses syntax-level errors, such as unknown identifiers resulting from misspellings or version-induced library shifts, by providing the model with semantically relevant Mathlib documentation. 
Second, the Reasoner performs high-level structural and strategic analysis to assist the Prover in resolving tactic-related failures (e.g., ``linarith failed to find a contradiction''). 
The ``hard'' mode presents a clear performance-resource trade-off: while it incurs higher latency and increased context overhead, it yields substantial gains in error-correction efficiency and proof reliability.
We further analyze this tradeoff in the ablation study.

\subsection{Main Results}
\textbf{Putnam 2025.} The Putnam is a math competition that assesses advanced problem-solving across algebra, analysis, combinatorics, and number theory.
In Putnam 2025, our agent exhibits strong effectiveness and exceptional efficiency in automated formal theorem proving. 
As shown in Table \ref{tab:putnam}, under a fixed budget, MechMath-Easy successfully solves 9 of the 12 problems, and MechMath-Hard solves 11, substantially outperforming other baselines on average in terms of solved problems across all four metrics.
MechMath-Easy merely requires less than \$10 and 70 minutes per question on average.
Intuitively, this efficiency stems from our sorrifier-driven formal decomposition workflow: the agent identifies errors precisely where they arise, avoiding repeated analysis and unnecessary auxiliary lemmas.
Although MechMath does not solve A5 within the \$100 limit, the performance profiles of Axiom and Numina confirm the high computational demands of this problem. 
Due to experimental constraints and a tight budget, no further attempts for A5 are performed.

\begin{table}[t]
    \footnotesize
    \centering
    \begin{minipage}[b]{.45\textwidth}
    \centering
    \begin{subtable}{\linewidth}
    \subcaption{Time (min)}
    \begin{tabular*}{.99\linewidth}{@{\hspace{2pt}\extracolsep{\fill}} l c c c c c @{\hspace{2pt}}}
        \toprule
         & P1 & P3 & P4 & P5 & Avg\\
        \toprule
        Numina & 277 & 132 & 360 & 324 & 273\\
        Seed & 990 & 300 & 480 & 60 & 458 \\
        \cmidrule(lr){1-6}
        \textbf{MM-Hard} & 363 & 137 & 178 & 110 & 197 \\
        \bottomrule
    \end{tabular*}
    \end{subtable}
    ~\\
    \begin{subtable}{\linewidth}
    \subcaption{Cost (\$)}
    \begin{tabular*}{.99\linewidth}{@{\hspace{5pt}\extracolsep{\fill}} l c c c c c@{\hspace{5pt}}}
        \toprule
         & P1 & P3 & P4 & P5 & Avg\\
        \toprule
        \textbf{MM-Hard} & 141 & 43 & 95 & 47 & 82\\
        \bottomrule
    \end{tabular*}
    \end{subtable}
    \caption{Quantitative results on the IMO 2025 benchmark. MechMath spends the lowest time on average.}
    \label{tab:imo}
    \end{minipage}
    \hfill
    \begin{minipage}[b]{.51\textwidth}
    \footnotesize
    \centering
    \begin{tikzpicture}
        \begin{axis}[
            width=7.7cm,
            height=4.2cm,
            axis lines=left,
            xlabel={Depth},
            ylabel={\#Nodes},
            xmin=0, xmax=13,
            ymin=0, ymax=50,
            xtick={1, 2, 3, 4, 5, 6, 7, 8, 9, 10, 11, 12},
            ytick={0, 8, 16, 24, 32, 40, 48},
            tick align=outside,
            legend style={draw=none,at={(1,1)}},
            xmajorgrids=false,
            ymajorgrids=true,
            grid style={dashed, gray!30},
        ];
    
        \addplot[
            color=blue,
            mark=triangle*,
            mark options={fill=blue},
            thick,
        ]
        coordinates {
            (1, 9)
            (2, 48)
            (3, 21)
            (4, 3)
            (5, 2)
        };

        \addplot[
            color=red,
            mark=*,
            mark options={fill=red},
            thick,
        ]
        coordinates {
            (1, 2)
            (2, 5)
            (3, 4)
            (4, 1)
            (5, 1)
            (6, 3)
            (7, 3)
            (8, 2)
            (9, 2)
            (10, 1)
            (11, 1)
        };

        \addplot[
            color=orange,
            mark=square*,
            mark options={fill=orange},
            thick,
        ]
        coordinates {
            (1, 2)
            (2, 7)
            (3, 8)
            (4, 7)
            (5, 10)
            (6, 13)
            (7, 13)
            (8, 7)
            (9, 6)
            (10, 11)
            (11, 4)
            (12, 1)
        };
        \legend{MechMath, Aristotle, Seed};
        
        \end{axis}
    \end{tikzpicture}
    \captionof{figure}{Statistics of proof trees of IMO 2025 P4. The tree of MechMath is wide but shallow, while others' are relatively deep.}
    \label{fig:imo_nodes}
    \end{minipage}
\end{table}

\textbf{IMO 2025.} 
We further test four problems from IMO 2025.
%
From Table~\ref{tab:imo}, MechMath-Hard only uses less than 200 minutes on average, which is roughly 2/3 time of Numina.
This gap widens further when compared with Seed, with our running time less than half of Seed’s.
Further analysis of the proof structures shows that our proof trees are broader but shallower, while those generated by other methods contain multiple deeply nested nodes, as shown in Figure~\ref{fig:imo_nodes} (refer to Appendix~\ref{app:tree} for detailed tree structures). 
Since proofs at the same level can be parallelized, whereas parent nodes must wait for their child nodes to be completed, our shallower structure contributes to greater efficiency.
Beyond time and monetary cost, our generated proofs are distinguished by their minimal average length and small number of auxiliary lemmas (Table~\ref{tab:putnam} and Table~\ref{tab:app_putnam}), reflecting a highly disciplined and focused proof style.
Refer to Appendix~\ref{app:eg_proof} for 
the formal proof of the main theorem of IMO 2025 P3 as an example and \ref{app:putnam_imo} for more results.

\textbf{MiniF2F.}
To further validate scalability and robustness, we evaluate our approach on the miniF2F benchmark (Table~\ref{tab:minif2f}). 
MechMath-Hard attains a 99.2\% pass rate (242/244), matching state-of-the-art (Hilbert, SeedProver), yet requires only 512 LLM calls versus Hilbert's 11.3K, substantially reducing inference cost. 
Moreover, our workflow produces shorter, more structured proofs. 
Figure~\ref{fig:minif2f_passrate} shows pass rates across inference budgets, with details in Appendix \ref{app:minif2f}.

\textbf{ProverBench.}
We further evaluate on ProverBench, following DreamProver's protocol on 40 number theory problems~\citep{zhang2026dreamprover}. 
As shown in Table~\ref{tab:proverbench}, MechMath solves 24 of 40 problems, outperforming Hilbert (Gemini 3.1 Pro Preview, 16 solved) and DeepSeek Prover and Goedel Prover ($<$12 with Pass@32). 
Notably, MechMath also identifies 4 erroneous problems (p6, p13, p20, p28) in the dataset. 
These results confirm our core findings and demonstrate MechMath's accuracy and efficiency across diverse benchmarks. 
Cost breakdown and case studies are provided in Appendix \ref{app:proverbench}.

\begin{table}[t]
\footnotesize
\centering
\begin{minipage}{.48\linewidth}
\centering
\begin{subtable}{\linewidth}
    \subcaption{Comparison of easy and hard mode}
    \begin{tabular*}{.99\linewidth}{@{\hspace{5pt}\extracolsep{\fill}} l c c c c @{\hspace{5pt}}}
        \toprule
         & A1 & A2 & A3 & A4 \\
        \toprule
        \multicolumn{5}{c}{Cost (\$)}\\
        \midrule
        Easy & 1.53 & 4.89 & 11.6 & 19.9 \\
        + search & 3.28 & 13.3 & 11.1 & 13.7 \\
        + comment & 6.33 & 5.70 & 6.19 & 6.42 \\
        + both (Hard) & 4.21 & 29.0 & 19.5 & 10.5 \\
        \toprule
        \multicolumn{5}{c}{\#Lemmas}\\
        \midrule
        Easy & 1 & 4 & 7 & 15 \\
        + search & 4 & 12 & 9 & 18 \\
        + comment & 8 & 9 & 9 & 7 \\
        + both (Hard) & 1 & 20 & 9 & 10 \\
        \bottomrule
    \end{tabular*}
\end{subtable}
~\\[15pt]
\begin{subtable}{\linewidth}
    \subcaption{Influence of the informal proof}
    \begin{tabular*}{.99\linewidth}{@{\hspace{5pt}\extracolsep{\fill}} l c c c c @{\hspace{5pt}}}
        \toprule
         & A1 & A2 & A3 & A4\\
        \toprule
        \multicolumn{5}{c}{w/ informal proof}\\
        \midrule
        Cost (\$) & 4.21 & 29.0 & 19.5 & 10.5 \\
        \#Nodes & 1 & 20 & 9 & 10\\
        \toprule
        \multicolumn{5}{c}{w/o informal proof}\\
        \midrule
        Cost (\$) & 3.90 & \ding{55} & 23.3 & 20.5 \\
        \#Nodes & 2 & \ding{55} & 11 & 44\\
        \bottomrule
    \end{tabular*}
\end{subtable}
\end{minipage}
\hfill
\begin{minipage}{.5\linewidth}
\centering
\begin{subtable}{\linewidth}
    \subcaption{Influence of the model}
    \begin{tabular*}{.99\linewidth}{@{\hspace{4pt}\extracolsep{\fill}} l c c c c @{\hspace{4pt}}}
        \toprule
         & A1 & A2 & A3 & A4\\
        \toprule
        Gemini 3.1 Pro & \checkmark & \checkmark & \checkmark & \checkmark \\
        Claude 4.5 Opus & \checkmark & \checkmark & \checkmark & \checkmark \\
        GPT-5.5 & \checkmark & \checkmark & \checkmark & \checkmark \\
        Deepseek 3.2 Reasoner & \checkmark & \ding{55} & \ding{55} & \ding{55} \\
        Seed 2 Pro & \checkmark & \ding{55} & \ding{55} & \ding{55} \\
        Kimi K 2.5 & \checkmark & \ding{55} & \ding{55} & \ding{55} \\
        \bottomrule
    \end{tabular*}
\end{subtable}
~\\
\begin{subtable}{\linewidth}
    \subcaption{Effectiveness of the Sorrifier mechanism}
    \begin{tabular*}{.99\linewidth}{@{\hspace{5pt}\extracolsep{\fill}} l c c c c @{\hspace{5pt}}}
        \toprule
        Metric & A1 & A2 & A3 & A4 \\
        \toprule
        \multicolumn{5}{c}{Discard \& Retry}\\
        \midrule
        LLM Calls & \ding{55} & \ding{55} & \ding{55} & \ding{55} \\
        Time (min) & \ding{55} & \ding{55} & \ding{55} & \ding{55} \\
        Cost (\$)  & \ding{55} & \ding{55} & \ding{55} & \ding{55} \\
        \toprule
        \multicolumn{5}{c}{Iterative Repair}\\
        \midrule
        LLM Calls & 4 & \ding{55} & 42 & 126 \\
        Time (min) & 20 & \ding{55} & 141 & 518 \\
        Cost (\$)  & 1.59 & \ding{55} & 16.30 & 55.70 \\
        \toprule
        \multicolumn{5}{c}{MechMath}\\
        \midrule
        LLM Calls & 3 & 31 & 69 & 88 \\
        Time (min) & 18 & 45 & 106 & 114 \\
        Cost (\$)  & 1.53 & 4.89 & 11.60 & 19.90 \\
        \bottomrule
    \end{tabular*}
\end{subtable}
\end{minipage}
\caption{Ablation study. (b), (c) and (d) is conducted on MechMath-Hard, since the easy mode fails in most of these settings.}
\label{tab:alation}
\end{table}

We provide more analysis in the appendix, including proof structure preservation (Appendix \ref{app:structural_preservation}), subgoal verification (Appendix \ref{app:global_flaw}) and Sorrifier mechanism (Appendix \ref{app:sorrifier_scope}).

\subsection{Ablation Study}

The ablation study encompasses four aspects: (1) comparison of the easy and the hard mode; (2) the necessity of the informal proving process; (3) the scalability of our pipeline across different models; and (4) the effectiveness of Sorrifier.
Due to budget constraints, the ablation experiments are conducted only on the first four Putnam problems.

\textbf{Easy/Hard modes.} 
To evaluate the performance-resource trade-off of incorporating Lean-specific feedback modules, we conducted an ablation study across three configurations: the addition of a search module, a comment module, and both of them (the ``hard mode''). 
For relatively straightforward problems (A1 and A2), the inclusion of auxiliary tools proved counterproductive, increasing costs and occasionally misleading the model's reasoning (Table~\ref{tab:alation}(a)). 
However, as problem complexity increased (A3 and A4), these tools became instrumental, effectively reducing both monetary cost and the volume of auxiliary lemmas. 
Furthermore, for problems B1 and B6, which remained intractable in the ``easy'' mode, the synergy of these modules enabled successful resolution, as detailed in Table~\ref{tab:putnam}.

\textbf{Informal proof.}
As shown in Table~\ref{tab:alation}(b), our agent benefits substantially from the integration of the informal proof.
Despite the fact that incorporating an informal proving step introduces additional overhead for the exceptionally simple problem A1, its benefits become evident as the difficulty of the problem increases. 
Specifically, a well-structured informal proof reduces errors in formal proofs, thereby decreasing the size of the proof tree and ultimately lowering the overall overhead. 
Moreover, for problem A2, the absence of an informal proof renders the prover incapable of completing the proof.
Motivated by the observation that an informal proof may be unnecessary for simple theorems, we introduce an additional fast proving step prior to the informal proving process to improve efficiency. 

\textbf{Different Models.}
To test the scalability of our pipeline across different models, we use various models as the reasoner and prover, including 
Gemini 3.1 Pro~\citep{google2026gemini31pro}, 
Claude 4.5 Opus~\citep{claude2025claude45}, 
GPT-5.5~\citep{openai2026gpt55}, 
Deepseek 3.2 Reasoner~\citep{deepseekai2025deepseekv32}, 
Seed 2.0 Pro~\citep{seed2026seed2pro}, 
and Kimi K2.5~\citep{kimi2026kimik25}.
As reported in Table~\ref{tab:alation}(c), with our formal decomposition pipeline, Gemini 3.1 Pro, Claude 4.5 Sonnet and GPT-5.5 can solve all four questions.
Constrained by the Lean capabilities, other three models manage to solve only one problem. 
Yet even this modest result reflects a substantial improvement over the original model, where no problems could be solved.


\textbf{Sorrifier}
To evaluate the efficacy of the Sorrifier mechanism, we conducted experiments on Putnam A1-A4 under a strict budget of 128 LLM calls, comparing it against two baselines:
(1) \textit{discard \& retry}: The system first generates an informal proof and subsequently performs up to 128 independent formal generation attempts.
(2) \textit{iterative repair}: Upon producing the initial formal proof, the system directly executes up to 127 consecutive repair iterations, with the corresponding error feedback provided in each round.
As summarized in Table~\ref{tab:alation}(d), the \textit{discard \& retry} strategy failed entirely (0/4), while the \textit{iterative repair} strategy solved only three problems (3/4). 
In contrast, although MechMath consumes more LLM calls on certain problems to extract and evaluate subgoals, it significantly reduces overall inference time and API costs. 
This confirms the Sorrifier mechanism's key benefit: localizing errors avoids full regeneration and context blow-up, yielding more efficient formal proof derivation.

\section{Conclusion}
In summary, we propose a novel agent workflow-\textbf{MechMath}, which adopts a strategy centered on sorrifier-driven formal decomposition and significantly improves efficiency. 
It novelly leverages the \texttt{sorry} placeholder to surgically isolate and resolve localized errors while preserving the structurally correct surrounding proof. 
This approach enables iterative subgoal decomposition without discarding valuable progress, effectively bridging the gap between informal reasoning and formal verification. 
Experimental results on challenging mathematical competition benchmarks, including IMO 2025 and Putnam 2025, demonstrate that MechMath achieves significant improvements in proving efficiency, highlighting the effectiveness of formal decomposition over traditional informal approaches.

\textbf{Limitations and Future Work.}
MechMath employs a fixed pipeline that follows an iterative cycle of informal proof, formal proof, and subgoal splitting. 
However, an ideal agent may require the flexibility to adapt based on context. 
For example, it can omit the informal proof stage for simpler problems or perform additional error-fix iterations for formal proof on harder ones. 
In future work, we will modularize the sorrifier as a standalone component integrated into the harness. 
This design preserves the efficacy of the sorrifier-driven formal decomposition while enabling the model to autonomously invoke the component as needed.

\clearpage
\section*{Acknowledgment}
This paper is supported by the Strategic Priority Research Program of CAS Grants XDA0480502 and XDA0480503.

\bibliography{ref}
\bibliographystyle{colm2026/colm2026_conference}

\clearpage
\appendix
\onecolumn
\section{Brief Introduction to Lean}
\label{app:lean}
Lean is an open-source interactive theorem prover (ITP) and a functional programming language developed primarily by \citet{moura2021lean}. 
Built upon a version of dependent type theory known as the \textit{Calculus of Inductive Constructions} (CiC), Lean provides a rigorous framework for formalizing mathematical definitions and verifying the correctness of proofs. 

Unlike traditional informal mathematics, Lean requires that every logical step be verified by a small, trusted kernel, ensuring a high degree of mathematical certainty. 
It has gained significant traction within the mathematical community, notably through the \textit{mathlib} project~\citep{mathlib}, which is a massive collaborative library of formalized mathematics. 
In practice, users employ various tactics to resolve logical goals.
The following table summarizes the primary commands (tactics) used within a \texttt{by} block to manipulate the proof state and construct formal arguments.

\begin{table}[h]
\centering
\scriptsize
\begin{tabular*}{\linewidth}
    {@{\hspace{5pt}\extracolsep{\fill}} l m{260pt} l @{\hspace{5pt}}}
\toprule
\textbf{Tactic} & \textbf{Functionality} & \textbf{Example Context} \\ \midrule
\texttt{have} & Introduces an intermediate assertion or lemma to the local context. & \texttt{have h : n > 0 := ...} \\
\texttt{let} & Defines a local term or function with a specific value. & \texttt{let f := n + 1} \\
\texttt{apply} & Matches the current goal with the conclusion of a theorem. & \texttt{apply Nat.add\_comm} \\
\texttt{exact} & Closes a goal that matches a given term or hypothesis precisely. & \texttt{exact h} \\
\texttt{intro} & Moves a universal quantifier ($\forall$) or premise ($\to$) into the context. & \texttt{intro n hn} \\
\texttt{cases} & Performs case analysis on a hypothesis or inductive type. & \texttt{cases h with | hl | hr} \\
\texttt{rw} & Substitutes terms in the goal using an equality or equivalence. & \texttt{rw [h\_eq]} \\
\texttt{simp} & Invokes a simplification engine to automate trivial rewrites. & \texttt{simp [add\_assoc]} \\ \bottomrule
\end{tabular*}
\caption{Fundamental Tactics for Formal Proof Construction in Lean.}
\end{table}

\section{Real Examples in the MechMath Pipeline}
All examples presented in the main text are toy examples.
In this section, we provide some real examples collected from our experiments, which may be better understood.

\subsection{Example of a Logical Error in Subgoal Extracion}
\label{app:logic_error}
Here is an example of a logical error in subgoal extraction, which indicates the necessity of evaluation.
A Lean error exists in the proof of the lemma, and is replaced with the red \texttt{sorry} block by the sorrifier.
However, as the verifier model points out, there is a logical flaw here, which results in an impossibility to prove.
This kind of subgoal should not be extracted.
Instead, the prover will try to fix the whole proof to ensure the logical soundness.
\small

\noindent\textbf{Part of the proof for a lemma:}


\begin{lstlisting}[frame = single, style=lean]
lemma putnam_2025_a2-0 (x : ℝ) (hx : x ∈ Icc (0 : ℝ) Real.pi) (h1 : sin x ^ (2 : ℕ) ≤ x ^ (2 : ℕ)) :
  (1 : ℝ) / Real.pi * x * (Real.pi - x) ≤ sin x := by
  have hpi_pos : 0 < Real.pi := Real.pi_pos
  have h_bound : ∀ y ∈ Icc (0 : ℝ) (Real.pi / 2), (1 : ℝ) / Real.pi * y * (Real.pi - y) ≤ sin y := by
    intro y hy
    have hy_nonneg : 0 ≤ y := hy.1
    have hy_half : y ≤ Real.pi / 2 := hy.2
    have h_jordan := Real.mul_le_sin hy_nonneg hy_half
    have h_quad_le : (1 : ℝ) / Real.pi * y * (Real.pi - y) ≤ (2 : ℝ) / Real.pi * y := by
      have hy_pi_sub : Real.pi - y ≤ Real.pi := by linarith [hy.1]
      calc (1 : ℝ) / Real.pi * y * (Real.pi - y)
        _ ≤ (1 : ℝ) / Real.pi * y * 2 := by
            `\redhl{sorry}`
        _ = (2 : ℝ) / Real.pi * y := by ring
    exact le_trans h_quad_le h_jordan
  rcases le_or_gt x (Real.pi / 2) with h_half | h_half
  · exact h_bound x ⟨hx.1, h_half⟩
  · have h_symm : sin (Real.pi - x) = sin x := Real.sin_pi_sub x
    have h_pi_sub : Real.pi - x ∈ Icc (0 : ℝ) (Real.pi / 2) :=
      ⟨by linarith [hx.2], by linarith [h_half]⟩
    calc (1 : ℝ) / Real.pi * x * (Real.pi - x)
      _ = (1 : ℝ) / Real.pi * (Real.pi - x) * (Real.pi - (Real.pi - x)) := by ring
      _ ≤ sin (Real.pi - x) := h_bound (Real.pi - x) h_pi_sub
      _ = sin x := h_symm
\end{lstlisting}

\noindent\textbf{Analysis by the verifier:}
The goal is to prove: $ \frac{1}{\pi} y (\pi - y) \le \frac{2}{\pi} y$. Let's test this inequality for a valid value of $y$, such as $y=0.5$, which clearly sits within $[0,\pi/2]$ (since $\pi/2\approx1.57$). Substituting $y=0.5$ into the left-hand side: $ \frac{1}{\pi} (0.5) (\pi - 0.5) = 0.5 - \frac{0.25}{\pi} \approx 0.5 - 0.079 = 0.421$. Substituting $y=0.5$ into the right-hand side: $ \frac{1}{\pi} (0.5) (2) = \frac{1}{\pi} \approx 0.318 $. The inequality evaluates to $0.421\leq0.318$, which is \redhl{mathematically strictly false}.

\subsection{Example of a Proof Written by MechMath}
\label{app:eg_proof}
\normalsize
The following Lean code is part of the proof of IMO 2025 P4, which is generated by MechMath.
The proof structure is highly disciplined, with each \texttt{have} block corresponding to a specific step in the informal sketch (marked \textcolor{blue}{blue}).
More importantly, instead of naively translating the informal sketch into formal proof, prover will fix the error according to the Lean feedback, Lean search results, and comments of verifier.
For example, the prover adds two more \texttt{have} blocks between \texttt{step6} and \texttt{step7} to supplement necessary logic.

Within the \texttt{have} block, some Lean proofs contain errors.
Through the formal decomposition workflow, these error blocks are first converted into \texttt{sorry} by the sorrifier, then extracted as subgoals for further proving.
Once all subgoals have been proven, they are assembled to the original proof (marked \textcolor{orange}{orange}).
Detailed proof of subgoals is omitted.
All proofs of our experiments can be accessed in the github repository.

\begin{lstlisting}[frame = single, style = lean]
theorem imo_2025_p3 (IsBonza : (ℕ+ → ℕ+) → Prop)
  (hisBonza : ∀ f, IsBonza f ↔ ∀ a b : ℕ+, (b : ℕ) ^ (a : ℕ) ≡ (f b : ℕ) ^ (f a : ℕ) [MOD f a]) :
  IsLeast {c : ℝ | ∀ f, IsBonza f → ∀ n : ℕ+, (f n : ℝ) ≤ c * (n : ℝ)} 4 := by
  constructor
  · intro f hf n
    have hBonza : ∀ a b : ℕ+, (b : ℕ) ^ (a : ℕ) ≡ (f b : ℕ) ^ (f a : ℕ) [MOD f a] := (hisBonza f).mp hf
    have `\bluehl{step1}` : f 1 = 1 := by exact `\orangehl{imo\_2025\_p3\_0}` f hBonza
    have `\bluehl{step2}` : ∀ p a : ℕ+, Nat.Prime (p : ℕ) → (p : ℕ) ∣ (f a : ℕ) → (p : ℕ) ∣ (a : ℕ) := by exact `\orangehl{imo\_2025\_p3\_1}` f hBonza step1
    have `\bluehl{step3}` : ∀ a : ℕ+, padicValNat 2 (f a : ℕ) ≤ padicValNat 2 (a : ℕ) + 2 := by
      exact fun (a : ℕ+) ↦ `\orangehl{imo\_2025\_p3\_2}` f hBonza step1 step2 a
    have `\bluehl{step4}` : (∃ p : ℕ+, Nat.Prime (p : ℕ) ∧ (p : ℕ) % 2 = 1 ∧ (f p : ℕ) > 1) → ∀ a : ℕ+, f a = a := by
      exact
        fun (a : ∃ (p : ℕ+), Nat.Prime ↑p ∧ ↑p % (2 : ℕ) = (1 : ℕ) ∧ ↑(f p) > (1 : ℕ)) (a_1 : ℕ+) ↦
        `\orangehl{imo\_2025\_p3\_3}` IsBonza hisBonza f hf hBonza step1 step2 step3 a a_1
    have `\bluehl{step5}` : (∀ p : ℕ+, Nat.Prime (p : ℕ) → (p : ℕ) % 2 = 1 → (f p : ℕ) = 1) → ∀ a : ℕ+, (f a : ℝ) ≤ 4 * (a : ℝ) := by
      exact
        fun (a : ∀ (p : ℕ+), Nat.Prime ↑p → ↑p % (2 : ℕ) = (1 : ℕ) → ↑(f p) = (1 : ℕ)) (a_1 : ℕ+) ↦
        `\orangehl{imo\_2025\_p3\_4}` f hBonza step1 step2 step3 step4 a a_1
    have step_final : (f n : ℝ) ≤ 4 * (n : ℝ) := by
      by_cases h_exists : ∃ p : ℕ+, Nat.Prime (p : ℕ) ∧ (p : ℕ) % 2 = 1 ∧ (f p : ℕ) > 1
      . have h_fa_a : f n = n := step4 h_exists n
        have h_fa_R : (f n : ℝ) = (n : ℝ) := by exact_mod_cast h_fa_a
        have hn_pos : 0 ≤ (n : ℝ) := by positivity
        linarith
      . push_neg at h_exists
        have h_all : ∀ p : ℕ+, Nat.Prime (p : ℕ) → (p : ℕ) % 2 = 1 → (f p : ℕ) = 1 := by
          exact fun (p : ℕ+) (a : Nat.Prime ↑p) (a_1 : ↑p % (2 : ℕ) = (1 : ℕ)) ↦
            `\orangehl{imo\_2025\_p3\_5}` f h_exists p a a_1
        exact step5 h_all n
    exact step_final
  · intro c hc
    let F_val (n : ℕ) : ℕ :=
      if n % 2 = 1 then 1
      else if n % 4 = 2 then 2
      else 4 * 2 ^ padicValNat 2 n
    have hF_pos : ∀ n : ℕ+, 0 < F_val (n : ℕ) := by
      intro n
      dsimp [F_val]
      split_ifs
      . omega
      . omega
      . positivity
    let F : ℕ+ → ℕ+ := fun n => ⟨F_val (n : ℕ), hF_pos n⟩
    have `\bluehl{step6}` : IsBonza F := by exact `\orangehl{imo\_2025\_p3\_6}` IsBonza hisBonza F_val rfl hF_pos F rfl
    have step7_F_val : F_val 4 = 16 := by exact `\orangehl{imo\_2025\_p3\_7}` F_val rfl
    have step7_F4 : (F (4 : ℕ+) : ℝ) = 16 := by
      have h4 : (F (4 : ℕ+) : ℕ) = 16 := step7_F_val
      exact_mod_cast h4
    have `\bluehl{step7}` : 4 ≤ c := by exact `\orangehl{imo\_2025\_p3\_8}` IsBonza c hc F step6 step7_F4
    exact step7
\end{lstlisting}

\subsection{Example of the Proof Tree}
Proof tree of the IMO 2025 P3 and P5. The innermost node is the root nodes. Outer small sectors of each sector is the child nodes of that node. The proof tree generated by MechMath is wide but relatively shallow.
\label{app:tree}

\begin{figure}[h]
    \centering
    \begin{subfigure}{.5\linewidth}
        \includegraphics[width=\linewidth, trim=5pt 100pt 120pt 140pt, clip]{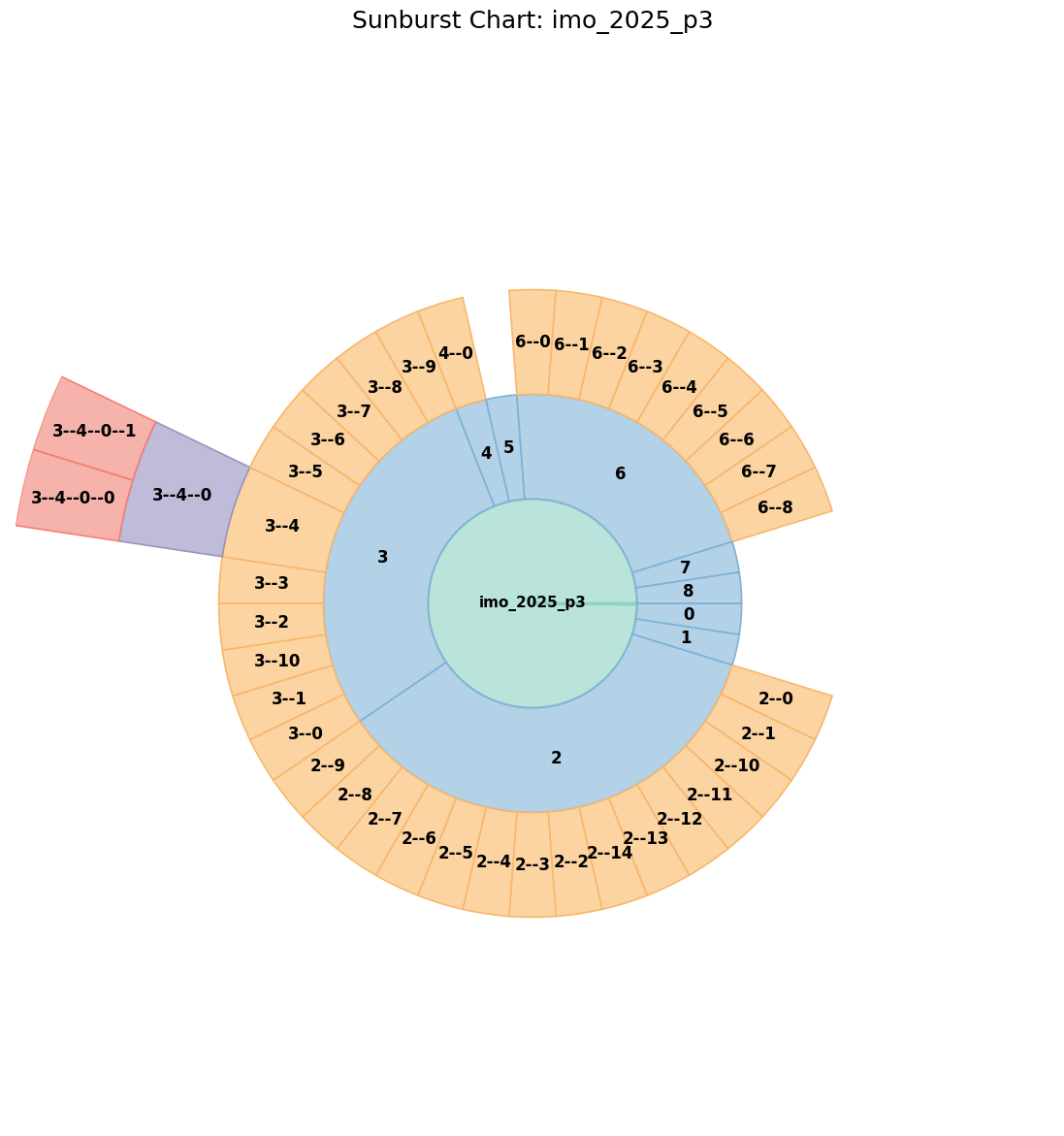}
        \subcaption{Tree of IMO 2025 P3}
    \end{subfigure}
    \hfill
    \begin{subfigure}{.45\linewidth}
        \includegraphics[width=\linewidth, trim=5pt 60pt 70pt 110pt, clip]{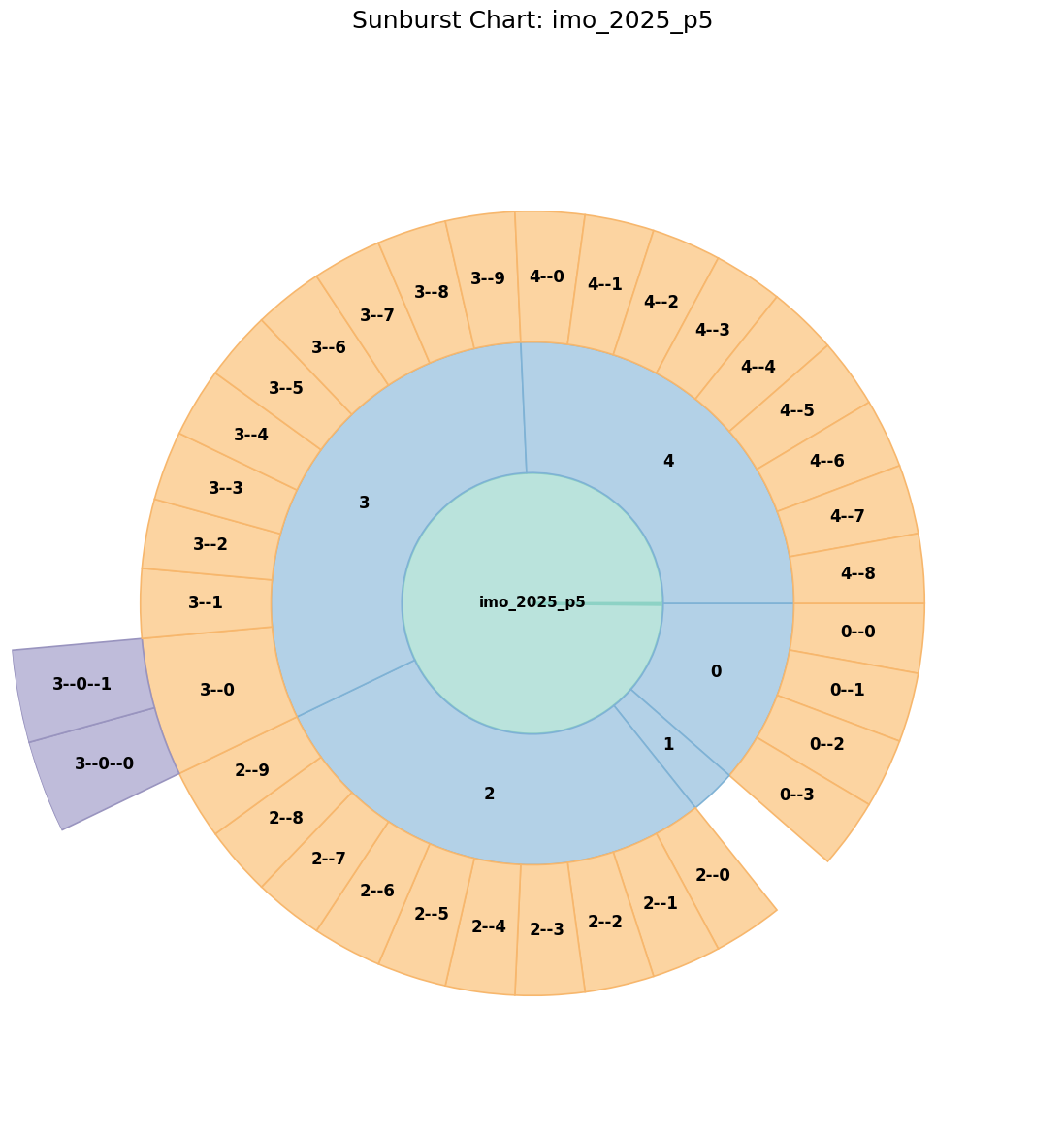}
        \subcaption{Tree of IMO 2025 P5}
    \end{subfigure}
    \caption{Proof tree of the IMO 2025 P3 and P5. The innermost node is the root nodes. Outer small sectors of each sector is the child nodes of that node. The proof tree generated by MechMath is wide but relatively shallow.}
    \label{fig:tree}
\end{figure}

\section{Implementation Details of Experiments}
\label{app-exp}
\subsection{Experiment Settings}
\textbf{Datasets.} 
To comprehensively evaluate the effectiveness of MechMath, we select Putnam 2025 and IMO 2025. William Lowell Putnam 2025 is a notoriously challenging mathematics competition for undergraduate students, containing 12 problems from different fields of mathematics. IMO 2025, the International Mathematical Olympiad, is the most prestigious high school mathematics competition, featuring 6 problems that test combinatorial, geometric, and number-theoretic insights. 
IMO 2025 Problem 2 is a geometric problem that is difficult to formalize in Lean and is usually solved with different approaches such as AlphaGeometry~\citep{Alphageometry}; therefore, we exclude it from the test. 

\textbf{Evaluation.}
To comprehensively evaluate our method's performance, we report the results on generating valid formal proofs for each problem, alongside the corresponding line counts and the number of lemmas. Furthermore, to assess efficiency, we also record the time expenditure and API cost for full proof generation. We assign a budget of \$100 for Putnam 2025 and \$200 for IMO, except \$300 for Putnam 2025 A5, which is widely recognized for its difficulty due to the absence of essential theorems. 

\textbf{Baselines.} We compare our results against several state-of-the-art baselines: Seed-Prover 1.5, Axiom, and Numina-Lean-Agent, based on their reported performance. In addition, we also compare with Hilbert, using Goedel Prover 32B as the formal prover and pairing it with Gemini 3.1 Pro Preview as the informal model, which is the same base model used for our agent, to maintain a fair and consistent experimental environment. For other hyperparameters, we rigorously set them to their experimental configuration. We also allocate the same budget for Hilbert to perform a fair comparison. 

\textbf{Implementation Details.} 
For the implementation of our MechMath, we use Gemini-3.1-Pro-Preview as the base model, whose training data, as mentioned in their model card, is cut off in January 2025, thereby avoiding data leakage. We use LeanDex as our retrieval engine to retrieve the relevant Mathlib theorems for the agent. For informal solution generation, we employ 16 iterations of refinement. In each iteration, we perform 3 verification passes to ensure the correctness of the generated proof. For the formal proving, we perform an initial formalization attempt, followed by up to 4 rounds of error correction based on Lean's compiler feedback. Subsequently, each error identified by the Sorrifier will undergo a 3-stage ``extract-assess'' evaluation. If a logical error is detected, the resulting diagnostic comments and the corresponding \texttt{sorry} locations are passed back to perform further error fixing. 
The prompts used during the process are provided in the next section.

\subsection{Algorithm of the Sorrifier}
\label{app:sorrifier}

\begin{algorithm}[H]
\caption{Sorrifier Algorithm: Iterative Proof Patching}
\label{algo:sorrifier}

\KwIn{Initial formal code $C$, Lean Compiler $Compiler$, Max iterations $N$}
\KwOut{Patched code $C_{final}$, Boolean $is\_success$}
\BlankLine

\For{$i \leftarrow 1$ \KwTo $N$}{
    \tcp{Cleanup redundant statements located after existing 'sorry'}
    $C \leftarrow \text{CleanUpRedundantCode}(C)$\;
    
    \tcp{Compile code and retrieve verification results}
    $result, errors \leftarrow Compiler.\text{Verify}(C)$\;
    
    \If{$result$ is Valid (allowing 'sorry')}{
        \Return{$(C, \text{True})$}\;
    }
    
    \If{$errors$ is empty \textbf{or} $Compiler$ timeouts}{
        \Return{$(C, \text{False})$}\;
    }
    
    \tcp{Locate the primary error's line and column}
    $err \leftarrow \text{SelectPrimaryError}(errors)$\;
    
    \tcp{Find all structural blocks (like 'have', 'calc') overlapping the error line}
    $overlapping\_blocks \leftarrow \text{FindBlocksContainingLine}(C, err.\text{line})$\;
    
    \tcp{KEY 1: Isolate the innermost block by finding the one with the minimum length}
    $innermost\_block \leftarrow \text{FindShortestBlock}(overlapping\_blocks)$\;
    
    \tcp{KEY 2: Apply different strategies based on whether the error is inside or outside a block}
    \eIf{$innermost\_block$ is Empty}{
        \tcp{Strategy for OUTSIDE: The error is not wrapped in a 'have' or similar block}
        \tcp{Truncate the statement at the error position and append 'sorry'}
        $C_{new} \leftarrow \text{TruncateLineAndAppendSorry}(C, err.\text{line}, err.\text{column})$\;
    }{
        \tcp{Strategy for INSIDE: The error occurs within a structured block}
        \eIf{$innermost\_block.\text{type} \in \{\text{calc}, \text{choose}\}$}{
            $C_{new} \leftarrow \text{ReplaceEntireBlockWithSorry}(C, innermost\_block)$\;
        }{
            \tcp{For 'have' or 'replace', target only the proof body}
            $C_{new} \leftarrow \text{ReplaceProofBodyWithSorry}(C, innermost\_block, err)$\;
        }
    }
    
    \tcp{Prevent infinite loops if no valid modification was made}
    \If{$C = C_{new}$}{
        \Return{$(C, \text{False})$}\;
    }
    
    $C \leftarrow C_{new}$\;
}
\Return{$(C, \text{False})$}\;
\end{algorithm}

\subsection{Code for Reproduction}
The code and materials required for reproduction are available at \url{https://github.com/MechMath/MechMath-v1/releases/tag/v1.0.0}.

\subsection{Prompts Used in the Experiment}
\subsubsection{Generate Informal Proof}
\ttfamily\small
You are a Formal Logic Expert and Mathematical Proof Engine. Your goal is to derive proofs that are rigorously structured, formalization-ready, and devoid of ambiguity.

Process Instruction: Before providing the final response, use your internal monologue/thinking process to fully solve the problem, verify every algebraic substitution, and ensure the logical chain is unbroken. Once your derivation is complete, you must output **exactly and only one $<$informal\_solution$>$$<$/informal\_solution$>$ block** containing the step-by-step proof (without thinking).

Core Constraints:

- Purely Algebraic/Symbolic: Do NOT use geometric intuition or visual symmetry. All geometric concepts must be translated into precise algebraic/analytic definitions.

- Atomic Steps: Decompose reasoning into the smallest possible logical units.

- No Hand-waving: Forbidden phrases include "obviously," "it is clear that," or "intuitively."

- Direct Output: Provide only the solution. Do NOT include internal monologue, "thought" blocks, or introductory/concluding conversational filler.

Formatting Requirements:

- Tags: Enclose the entire response in $<$informal\_solution$>$ tags.

- Structure: Present the proof as a discrete list of points.

- Naming Convention: In your final <informal\_solution> block, use the exact prefix Step X: (e.g., Step 1:, Step 2:) for each point, preceded by a bullet point (dot).

Derivation Instructions:

- Definitions: Explicitly state all variable types, definitions, and assumptions at the start.

- Step-by-Step Derivation: Number every step sequentially.

- Explicit Justification: For EACH step, explicitly state the rule of inference, algebraic identity, axiom, or theorem used.

- Calculations: Show every intermediate stage of simplification or substitution. Do not skip algebraic manipulation steps.

- No Formal Code: Focus solely on the informal solution of the problem. Do not write any Lean code.

Problem Statement: \\
\{problem\}

\subsubsection{Verify Informal Proof}
Your task is to evaluate the quality of a solution to a problem. The problem may ask for a proof of a statement, or ask for an answer. If finding an answer is required, the solution should present the answer, and it should also be a rigorous proof of that answer being valid.

Please evaluate the solution and score it according to the following criteria:

- If the solution is completely correct, with all steps executed properly and clearly demonstrated, then the score is 1

- If the solution is generally correct, but with some details omitted or minor errors, then the score is 0.5

- If the solution does not actually address the required problem, contains fatal errors, or has severe omissions, then the score is 0

- Additionally, referencing anything from any paper does not save the need to prove the reference. It's okay IF AND ONLY IF the solution also presents a valid proof of the reference argument(s); otherwise, if the solution omits the proof or if the proof provided is not completely correct, the solution should be scored according to the criteria above, and definitely not with a score of 1

Please carefully reason out and analyze the quality of the solution below, and in your final response present a detailed evaluation of the solution's quality followed by your score.

Therefore, your response should be in the following format:

Here is my evaluation of the solution:

[Your evaluation here. You are required to present in detail the key steps of the solution or the steps for which you had doubts regarding their correctness, and explicitly analyze whether each step is accurate: for correct steps, explain why you initially doubted their correctness and why they are indeed correct; for erroneous steps, explain the reason for the error and the impact of that error on the solution.]

Based on my evaluation, the final overall score should be: \textbackslash boxed{{...}}

[where ... should be the final overall score (0, 0.5, or 1, and nothing else) based on the above criteria]

Do not evaluate any semantics in informal solution related to Lean4; focus solely on the quality of the informal solution. If the informal solution mentions any specific Lean code, clearly point out that it should not include Lean code, and ignore the code in the evaluation.
---

Here is your task input:

\#\# Problem\\
\{problem\}

\#\# Solution\\
\{student\_solution\}

\subsubsection{Refine Informal Proof}
You are given a mathematical problem, an existing solution, and feedback on that solution.

Your task is to produce a **revised solution** that is more complete, rigorous, and clearly justified.

---

\#\#\# Problem\\
\{problem\}

---

\#\#\# Previous Solution\\
\{solution\}

---

\#\#\# Feedback\\
\{feedback\}

---

\#\#\# Instructions

- Process Instruction: Before providing the final response, use your internal monologue/thinking process to fully solve the problem, verify every algebraic substitution, and ensure the logical chain is unbroken. Once your derivation is complete, you must output **exactly and only one $<$informal\_solution$>$$<$/informal\_solution$>$ block** containing the step-by-step proof (without the thinking process).

- Carefully read the feedback and determine which points are **valid** and which may be due to **misunderstanding or evaluator error**.

- If you **agree** with a feedback item:

  - Revise the solution to fix the issue.
  
  - Add missing steps, clarify logical transitions, or strengthen rigor as needed.
  
- If you **disagree** with a feedback item:

  - Keep the original reasoning if it is correct.
  
  - Add **explicit explanations or clarifications** to prevent future misunderstandings.
  
- Do **not** simply restate the feedback.

- The final solution should be:

  - Self-contained
  
  - Logically coherent
  
  - Mathematically rigorous
  
  - Easy to follow for a careful reader
  
- Focus solely on the informal solution of the problem. DO NOT write any Lean code.
---

\#\#\# Output Format

- Use your internal monologue/thinking process to fully solve the problem, and provide **only** the revised solution in your final output.

- Provide exactly and only one $<$informal\_solution$>$ block.

- In your final $<$informal\_solution$>$ block, use the exact prefix Step X: (e.g., Step 1:, Step 2:) for each point, preceded by a bullet point (dot).

\subsubsection{System Prompt of Formal Proving and Fixing}
You are a Lean 4 and Mathlib expert.

Your task is to produce a complete, correct, and fully typechecking proof of a Lean 4 theorem.

\#\#\# Mandatory Rules (Must Follow Exactly)

1. Do NOT modify the formal statement in any way.\\
   - Do not change names, arguments, universes, implicit arguments, or formatting.\\
   - The theorem header must appear exactly as given in formal statement.

2. The final proof MUST be complete.\\
   - The proof must not contain \textasciigrave sorry\textasciigrave , \textasciigrave admit\textasciigrave , \textasciigrave by\_cases?\textasciigrave , \textasciigrave by\_contra?\textasciigrave , or any placeholder.\\
   - You CANNOT use the tactic \textasciigrave native\_decide\textasciigrave  and \textasciigrave aesop\textasciigrave .\\
   - All goals must be fully resolved.

3. Use only Lean 4 syntax compatible with Mathlib.\\
   - All identifiers must exist in Mathlib.\\
   - The code should compile in a standard Mathlib environment.\\
   - If a lemma from \textasciigrave search\_results\textasciigrave  is used, apply it with correct arguments.

\#\#\# Lean Hints:

Followings are some useful hints for Lean 4:

1. When dealing with inequalities, equalities and arithmetic operations like subtraction or division in \textasciigrave $\mathbb{N}$\textasciigrave  (natural numbers), beware of truncation. Use \textasciigrave $\mathbb{R}$\textasciigrave , \textasciigrave $\mathbb{Q}$\textasciigrave  or \textasciigrave $\mathbb{Z}$\textasciigrave  when possible for arithmetic operations. Avoid using \textasciigrave $\mathbb{N}$\textasciigrave  unless required by the theorem statement.\\
2. Be ESPECIALLY careful about implicit types while defining numeric literals. AVOID patterns like \textasciigrave 0 - 1\textasciigrave  or \textasciigrave 1 / 2\textasciigrave  without specifying the types.\\
3. ALWAYS specify types when dealing with numeric values to avoid ambiguities and unexpected behavior.\\
4. Use \textasciigrave simp only [specific\_lemmas]\textasciigrave  rather than bare \textasciigrave simp\textasciigrave  to avoid unpredictable simplifications.\\
5. Use \textasciigrave rw [$\rightarrow$ lemma]\textasciigrave  for reverse direction. When \textasciigrave rw\textasciigrave  fails, try \textasciigrave conv => rhs; rw [lemma]\textasciigrave  to target specific subterms. nth\_rw n [lemma] to rewrite only the nth occurrence.\\
6. When \textasciigrave ring\textasciigrave  fails on ring expressions, try \textasciigrave ring\_nf\textasciigrave  first to normalize, or cast to a concrete ring like \textasciigrave $\mathbb{R}$\textasciigrave  where the tactic works better.\\
7. Apply \textasciigrave norm\_num\textasciigrave  for concrete arithmetic calculations and \textasciigrave norm\_cast\textasciigrave  to simplify type coercions systematically.\\
8. Use \textasciigrave by\_contra h\textasciigrave  for proof by contradiction, which introduces the negation of the goal as hypothesis \textasciigrave h\textasciigrave .\\
9. If you get a \textasciigrave no goals to be solved\textasciigrave  error, it means that the previous tactics already solved the goal, and you can remove the subsequent tactics.\\
10. When proving theorems, ALWAYS write the proof in tactic mode, starting the proof with \textasciigrave := by\textasciigrave .\\
11. Do NOT use \textasciigrave begin\textasciigrave , \textasciigrave end\textasciigrave  blocks in your proof. This is invalid in Lean 4.\\
12. Use the \textasciigrave \textbackslash in\textasciigrave  rather than \textasciigrave in\textasciigrave  for any set (include for \textasciigrave \textbackslash sum\textasciigrave )\\
13. When working with inequalities or \textasciigrave calc\textasciigrave  blocks, pay special attention to the direction of the inequality signs.

\#\#\# Strategy Guidelines (Non-output)

- You may freely combine tactics (\textasciigrave simp\textasciigrave , \textasciigrave linarith\textasciigrave , \textasciigrave ring\textasciigrave , \textasciigrave nlinarith\textasciigrave , \textasciigrave omega\textasciigrave , \textasciigrave aesop\textasciigrave , etc.) and term-style proofs.\\
- You may introduce local \textasciigrave have\textasciigrave  or \textasciigrave let\textasciigrave  bindings inside the proof.\\
- Prefer robustness and clarity over cleverness.

\#\#\# Output

- You must return EXACTLY ONE \textasciigrave \textasciigrave \textasciigrave lean4\textasciigrave \textasciigrave \textasciigrave  code block.\\
- The code block must contain the final complete proof only.\\
- You must NOT include any other \textasciigrave \textasciigrave \textasciigrave lean4\textasciigrave \textasciigrave \textasciigrave  blocks anywhere in the output.

\subsubsection{Generate Formal Proof}
You need to formalize the provided `Informal Solution' to prove the provided `Formal Statement'. You MUST follow the structure of the informal solution closely, ensuring that each step in the informal solution corresponds to a clear and logically sound step in the Lean proof. Long and complete proof is expected for higher reward.

\#\#\# Formal Statement
\textasciigrave \textasciigrave \textasciigrave lean4
\{formal\_statement\}
\textasciigrave \textasciigrave \textasciigrave 

\#\#\# Informal Solution (MUST follow this structure)
\{informal\_solution\}

\#\#\# Critical Proof Principles

1. **No Reliance on "Ready-Made" Theorems**: \\
   - These problems are mathematically nontrivial. Do NOT solve the goal by a single \textasciigrave apply\textasciigrave of a powerful library lemma that encapsulates the entire result.\\
   - Use only basic definitions and elementary lemmas (e.g., rewriting definitions, arithmetic manipulation, manual inequalities).

2. **From First Principles \& Step-by-Step**:\\
   - Break the proof into small, explicit steps using \textasciigrave have\textasciigrave, \textasciigrave calc\textasciigrave, or intermediate lemmas.\\
   - Each formal step must correspond to a clear piece of reasoning. If a conclusion requires multiple ideas, split it.

3. **Proof Format**:\\
   - You CAN NOT write new lemma or definition in the \textasciigrave \textasciigrave \textasciigrave lean4\textasciigrave \textasciigrave \textasciigrave blocks.\\
   - You CAN NOT change the original formal statement.\\
   - Only ONE theorem or lemma is allowed in the lean block.\\
   - The proof must not contain \textasciigrave sorry\textasciigrave, \textasciigrave admit\textasciigrave, \textasciigrave by\_cases?\textasciigrave, \textasciigrave by\_contra?\textasciigrave, or any placeholder.

4. **Structure Alignment**:\\
   - The Lean proof must strictly mirror the structure of the provided Informal Solution. Do not introduce a fundamentally different strategy.\\
   - You MUST completely trust that the Informal Solution represents the most streamlined proof. Follow its proof method and process unconditionally, translating it strictly into Lean form without deviation.\\
   - For each step in the Informal Solution, formalize it into Lean code, preserving both the logical flow and mathematical reasoning of the original proof.

\subsubsection{Fix Formal Proof}
The Lean code you wrote before contains errors. Your task is to modify the code so that it compiles successfully, while preserving the original mathematical intent as much as possible. \\

\#\#\# Error Message from Lean:\\
\{error\_message\}

\{llm\_comments\}

\{search\_results\}

\#\#\# Critical Proof Principles

1. **No Reliance on "Ready-Made" Theorems**: \\
   - These problems are mathematically nontrivial. Do NOT solve the goal by a single \textasciigrave apply\textasciigrave of a powerful library lemma that encapsulates the entire result.\\
   - Use only basic definitions and elementary lemmas (e.g., rewriting definitions, arithmetic manipulation, manual inequalities).

2. **From First Principles \& Step-by-Step**:\\
   - Break the proof into small, explicit steps using \textasciigrave have\textasciigrave, \textasciigrave calc\textasciigrave, or intermediate lemmas.\\
   - Each formal step must correspond to a clear piece of reasoning. If a conclusion requires multiple ideas, split it.

3. **Proof Format**:\\
   - You CAN NOT write new lemma or definition in the \textasciigrave\textasciigrave\textasciigrave lean4\textasciigrave\textasciigrave\textasciigrave blocks.\\
   - You CAN NOT change the original formal statement.\\
   - Only ONE theorem or lemma is allowed in the lean block.\\
   - Make sure to follow all the rules and guidelines provided in the system prompt.

4. **Structure Alignment**:\\
   - Follow the structure of the original proof as closely as possible. Do not introduce a fundamentally different strategy.\\
   - Do not skip the steps included in the original proof. Instead, try to fix the errors while keeping the overall proof structure intact.

\#\#\# Hints for Fixing Errors:\\
- Pay attention to the mistakes made in history, avoid repeating them.\\
- When you find that a certain mistake cannot be corrected multiple times, try to change your approach.

\subsubsection{Comment on Formal Proof}
**Role**: You are an expert Lean 4 formal verification engineer and mathematician.

**Task**: Review the Lean 4 formal proof provided below (written by Claude) and provide strategic feedback.

**Rules**:\\
1. Scope: Do not comment on the problem statement or the hypothesis of the theorem itself. Focus entirely on the proof script.\\
2. Logic \& Dead Ends: Identify if the proof has entered a ``dead end'' (e.g., goals that are logically unprovable, circular reasoning, or contradictory states).\\
3. Proof Strategy: Focus on the high-level logic. Suggest more efficient or elegant proof paths. Provide a brief Lean snippet to demonstrate any recommended strategy.\\
4. Structural Analysis: Identify structural flaws in the proof (e.g., poor sub-goal management, deeply nested tactics, or redundant branching). Suggest better structures using \textasciigrave have\textasciigrave blocks where appropriate.\\
5. Code Snippets: Use brief Lean 4 code blocks (max 5-10 lines) to demonstrate the most important parts of the refactoring proof.\\
6. Tactic Tolerance: Do not nitpick specific Mathlib tactic names or minor syntax errors unless they fundamentally break the logic.

**Output Format**:\\
1. Numbered List Only: Use a numbered list (1, 2, 3...) for your comments.\\
2. Conciseness: Keep your feedback brief and technical.\\
3. No Conversational Filler: Do NOT include any introductory text, concluding remarks, or offers for further help (e.g., do not say ``Would you like me to...'' or ``I hope this helps'').\\
4. Strict Termination: Your response must end immediately after the final numbered point.\\
5. No comments on the problem statement or the hypothesis of the theorem itself.\\
6. You are preferred to use the code block in your comments for the important parts of the refactoring proof.

**Autual Task**:\\
Comment on the following Lean 4 formal proof:\\
\textasciigrave\textasciigrave\textasciigrave lean4\\
\{lean\_code\}\\
\textasciigrave\textasciigrave\textasciigrave
\normalfont\normalsize

\section{Full Experimental Results on multiple benchmarks}

\label{app:full_results}

In this section, we provide extended details and a more granular perspective on the miniF2F, ProverBench, Putnam, and IMO evaluation, including the LLM calls, token usage, time consumption, cost, lemma counts and proof length.

\subsection{Putnam 2025 and IMO 2025}
\label{app:putnam_imo}
To evaluate the capability of MechMath on more challenging mathematical problems, we subject the agent to elite-level competitions, specifically the Putnam 2025 and IMO 2025. 
Table~\ref{tab:putnam_metrics} and Table~\ref{tab:imo_metrics} summarize the problem-by-problem resource consumption across both benchmarks, documenting the total token usage, the total number of interactive environment recompilations, and the total volume of LLM calls. 
Compared to miniF2F and ProverBench, these advanced undergraduate and international olympiad problems exhibit a substantially higher baseline of structural difficulty, naturally eliciting scaling behavior from our iterative feedback loops. 
Specifically, for the Putnam 2025 benchmark, MechMath requires an average of 91.3 LLM calls and 1\,M tokens per problem. 
For the more formidable IMO 2025 dataset, the resource demand intensifies further, requiring an average of 643.5 LLM calls alongside approximately 8\,M tokens.

\begin{table}[ht]
    \centering
    \scriptsize
    \begin{tabular*}{\linewidth}
    {@{\hspace{5pt}\extracolsep{\fill}} l r r r r r r r r r r r r r @{\hspace{5pt}}}
        \toprule
        & A1 & A2 & A3 & A4 & A5 & A6 & B1 & B2 & B3 & B4 & B5 & B6 & Avg \\
        \midrule
        \#Tokens (K)  & 175 & 531 & 1,209 & 1,912 & - & 836 & 1,660 & 663 & 898 & 871 & 2,732 & 484 & 1,088 \\
        \#Sorrify & 2 & 8 & 16 & 16 & - & 8 & 24 & 5 & 10 & 9 & 28 & 13 & 12.6 \\
        LLM Calls    & 7 & 35 & 73 & 92 & - & 41 & 143 & 59 & 110 & 85 & 302 & 57 & 91.3 \\
        \bottomrule
    \end{tabular*}
    \caption{Detailed inference costs and proof metrics across the problems resolved by MechMath on Putnam 2025.}
    \label{tab:putnam_metrics}
\end{table}

\begin{table}[ht]
    \centering
    \footnotesize
    \begin{tabular*}{\linewidth}
    {@{\hspace{20pt}\extracolsep{\fill}} l r r r r r @{\hspace{20pt}}}
        \toprule
        & P1 & P3 & P4 & P5 & Avg \\
        \midrule
        \#Tokens (M)  & 13.77 & 4.81 & 8.34 & 5.04 & 7.99 \\
        \#Sorrify & 155 & 60 & 104 & 73 & 98.0 \\
        LLM Calls    & 981 & 450 & 726 & 417 & 643.5 \\
        \bottomrule
    \end{tabular*}
    \caption{Detailed inference costs and proof metrics across the problems resolved by MechMath on IMO 2025.}
    \label{tab:imo_metrics}
\end{table}

\begin{table*}[h]
    \scriptsize
    \centering
    \begin{tabular*}{\linewidth}
    {@{\hspace{5pt}\extracolsep{\fill}} l c c c c c c c c c c c c c @{\hspace{5pt}}}
        \toprule
         & A1 & A2 & A3 & A4 & A5 & A6 & B1 & B2 & B3 & B4 & B5 & B6 & Avg\\
        \toprule
        Hilbert & - & - & - & - & - & - & - & - & 1220 & - & - & - & 1220\\
        Seed & 631 & 469 & 927 & 1095 & - & 881 & 849 & 1613 & 584 & 628 & 2499 & 2594 & 1161\\
        Axiom & 556 & 458 & 1089 & 825 & 1878 & 468 & 1179 & 346 & 302 & 993 & 1310 & 862 & 763\\
        Numina & 365 & 401 & 422 & 605 & 3263 & 835 & 328 & 690 & 292 & 648 & 929 & 1820 & 667\\
        \cmidrule(lr){1-14}
        \textbf{MechMath-Easy} & 339 & 643 & 616 & 792 & - & 507 & - & 519 & 533 & 700 & 2022 & - & 741\\
        \textbf{MechMath-Hard} & \textbf{112} & 798 & 560 & 793 & - & \textbf{280} & 1425 & \textbf{224} & 627 & \textbf{568} & 1256 & \textbf{578} & \textbf{656}\\
        \bottomrule
    \end{tabular*}
    \caption{Proof length of the Putnam 2025 benchmark. The averages of Axiom and Numina are computed exclusive of A5.}
    \label{tab:app_putnam}
\end{table*}

\begin{table}[t]
    \footnotesize
    \centering
    \begin{subtable}{\linewidth}
        \centering
        \caption{Evaluation results on the miniF2F benchmark.}
        \label{tab:minif2f}
        \begin{tabular*}{.99\linewidth}{@{\hspace{10pt}\extracolsep{\fill}} l c l @{\hspace{10pt}}}
            \toprule
            Model & Acc & Note\\
            \midrule
            STP & 67.6 & Pass@25600\\
            Kimina-Prover-8B & 78.3 & Pass@32 \\
            Kimina-Prover-72B & 92.2 & TTRL \\
            DeltaProver & 95.9 & Pass@16384\\
            SeedProver & 99.6 & -\\
            BFS-Prover-V2-32B & 95.1 & w/ Planner\\
            Goedel-Prover-V2-8B & 90.2 & Pass@8192\\
            Goedel-Prover-V2-32B & 92.6 & Self-Correction\\
            Hilbert (GD) & 99.2 & 11.3K LLM Calls \\
            DeepSeek-Prover-V2-7B & 82.0 & Pass@8192 \\
            DeepSeek-Prover-V2-671B & 88.9 & Pass@8192\\
            Hilbert (DS) & 98.4 & Pass@8192 \\
            \cmidrule(lr){1-3}
            \textbf{MechMath-Hard (Ours)} & 99.2 & \makecell[l]{max: 512 LLM Calls\\avg: 57.6 LLM Calls}\\
            \bottomrule
        \end{tabular*}
    \end{subtable}
    ~\\
    \begin{subtable}{\linewidth}
        \centering
        \caption{Evaluation results on the number theory subset (40 problems) in ProverBench. Notably, MechMath successfully identifies 4 erroneous problems within the original dataset.}
        \label{tab:proverbench}
        \begin{tabular*}{.99\linewidth}{@{\hspace{10pt}\extracolsep{\fill}} l c l @{\hspace{10pt}}}
            \toprule
            Model & Acc & Note\\
            \midrule
            DeepSeek-Prover-V2-7B & 25.0 & Pass@32 \\
            Goedel-Prover-V2-8B & 25.0 & Pass@32 \\
            Goedel-Prover-V2-32B & 30.0 & Pass@32 \\
            Hilbert (GD) & 40.0 & Gemini 3.1 Pro \\
            \cmidrule(lr){1-3}
            \textbf{MechMath-Hard (Ours)} & 60.0 & \makecell[l]{max: 216 LLM Calls\\avg: 45.2 LLM Calls}\\
            \bottomrule
        \end{tabular*}
    \end{subtable}
    \caption{Pass rates on miniF2F and ProverBench-NT. MechMath achieves highly competitive performance with extremely low cost.}
    \label{tab:minif2f_proverbench}
\end{table}
\subsection{MiniF2F}
\label{app:minif2f}
Out of the 244 problems, this workflow successfully resolves 242, achieving an impressive 99.2\% overall pass rate.
Since some problems in miniF2F can be solved directly by Prover in one turn without relying on any informal solution, we exclude them from our data statistics in the following figures and tables, and focus only on the more challenging ones.

Figure~\ref{fig:minif2f_passrate} and Table~\ref{tab:minif2f_full} detail the performance metrics of the agent on miniF2F across four distinct inference-time dimensions. As illustrated, MechMath demonstrates a massive advantage in reducing computational overhead: to reach its peak pass rate, the agent utilizes a maximum of only 512 LLM calls and 4.73M tokens per problem—representing a 22.1$\times$ and 10.3$\times$ reduction compared to the 11.3K calls and 48.7M tokens demanded by Hilbert, respectively. The detailed statistics further reveal that our typical operational workload is vastly lighter. Among the successfully solved problems, MechMath requires an average of merely 57.6 LLM calls and 549K tokens. This exceptional efficiency in token and API usage is directly attributed to our systematic workflow: MechMath extracts each failed subproblem into a clean, self-contained context and resolves it independently. This strategy elegantly avoids both the computational waste of full proof regeneration and the excessive context length typically induced by repeated repairs.

This drastic reduction in computational volume translates directly into highly practical monetary costs: even under our maximum budget constraints, the cost is capped at \$42.47 per problem, with the actual execution average remaining as low as \$4.66. More importantly, regarding absolute time expenditure, the significant acceleration of MechMath is driven not only by the reduced API volume but also by our generated proof structure. Because proofs at the same level can be parallelized—whereas parent nodes in deeply nested structures must sequentially wait for their child nodes to be completed—our shallower proof trees contribute to vastly greater time efficiency, solving a typical problem in approximately 62 minutes (capped at 5.9 hours). This structural conciseness is further evidenced by the proof metrics: successfully generated solutions require an average of only 4.9 lemmas and 232.4 Lean lines. Ultimately, by combining the low-token isolation mechanism with a highly parallelizable shallow architecture, MechMath inherently bypasses the severe time latency and prohibitive API costs associated with the heavy sampling strategies of prior state-of-the-art baselines, yielding proofs that are both exceptionally efficient and highly disciplined.

\begin{table}[t]
    \centering
    \footnotesize
    \begin{tabular*}{\linewidth}
    {@{\hspace{10pt}\extracolsep{\fill}} l r r r r r r @{\hspace{10pt}}}
        \toprule
         & \# LLM Calls & \# Total Tokens & Time (min) & Cost (USD) & \# Lemmas & \# Lean lines \\
        \midrule
        Avg & 57.6 & 549\,K  & 62.1  & 4.66  & 4.9 & 232.4 \\
        Min & 5    & 35\,K   & 4.2   & 0.10  & 1   & 21    \\
        Max & 512  & 4.73\,M & 353.9 & 42.47 & 36  & 1442  \\
        \bottomrule
    \end{tabular*}
    \caption{Detailed inference costs and proof metrics across the problems resolved by MechMath on miniF2F (exclude simple prombles).}
    \label{tab:minif2f_full}
\end{table}

\input{app/pass_rate_budget}

\subsection{ProverBench}
\label{app:proverbench}

Across evaluations on the 40-theorem number theory subset of ProverBench and the Hilbert subset, MechMath successfully resolved 24/40 problems, significantly outperforming Hilbert's baseline of 16/40. Furthermore, MechMath demonstrates exceptional resource efficiency and operational control: resolving a target theorem requires an average of only 45.2 LLM calls (approximately 336K tokens) at a highly economical cost of \$2.97, yielding concise proof structures that average just 4.25 auxiliary lemmas and 166 lines of formal code. Even under the most complex scenarios, the system successfully resolves the problem within an upper bound of 216 calls and 1.62M tokens (\$14.89), thanks to our Sorrifier-based isolation mechanism that effectively prevents exponential context growth. This underscores both the effectiveness and efficiency of our proposed framework.

Notably, we also identified four problematic theorems within the benchmark, detailed as follows:
\begin{itemize}[leftmargin=*]
    \item \texttt{number\_theory\_p6}: The theorem attempts to formalize rational solutions for the Pell-like equation $x^2 - dy^2 = 1$ parameterized by $t \in \mathbb{Q}$. However, it neglects the constraint where the denominator $d \cdot t^2 - 1 = 0$. In Lean 4, division by zero is syntactically defined but mathematically breaks the intended logic for certain values of $t$ (e.g., when $d=1, t=1$). MechMath correctly flagged this as a failure in the stage of informal proof.
    \item \texttt{number\_theory\_p13}: This theorem states that all odd divisors $d$ of $5x^2 + 1$ are congruent to $1, 3, 7,$ or $9 \pmod{20}$. The formalization uses the integer type $\mathbb{Z}$ but fails to constrain $d$ to positive integers ($d > 0$). Consequently, negative odd divisors like $d = -1$ yield negative remainders under Lean's \% operator, which falsifies the disjunction and renders the theorem unprovable without an explicit positivity constraint.
    \item \texttt{number\_theory\_p20}: While the natural language docstring claims that no positive integer $n$ can be expressed as the product of six consecutive positive integers, the formal statement defines the variables using the natural number type $\mathbb{N}$ without enforcing $a > 0$ or $n > 0$. This introduces a trivial counterexample at the boundary: $a=0$ and $n=0$, where $0 \times 1 \times 2 \times 3 \times 4 \times 5 = 0^5 = 0$. Since a valid counterexample exists, the theorem's goal to prove False is inherently falsified.
    \item \texttt{number\_theory\_p28}: The theorem incorrectly asserts that for a prime $p \equiv 3 \pmod 4$, the remainder of $N = \prod_{k=0}^{p-2}(k^2+1)$ divided by $p$ is $4$. This is mathematically impossible; the true remainder is $2$. For instance, if $p = 3$, then $N = (0^2+1)(1^2+1) = 2$, and $2 \pmod 3 = 2 \neq 4$. The informal verifier consistently rejected all translation attempts for this problem due to this structural contradiction.
\end{itemize}

\begin{table}[t]
    \centering
    \footnotesize
    \begin{tabular*}{\linewidth}
    {@{\hspace{20pt}\extracolsep{\fill}} l r r r r r r @{\hspace{20pt}}}
        \toprule
         & \# LLM Calls & \# Total Tokens & Time (min) & Cost (USD) & \# Lemmas & \# Lean lines \\
        \midrule
            Avg & 45.2 & 336\,K  & 28.8  & 2.97  & 4.25 & 166 \\
            Min & 5    & 35\,K   & 5.7   & 0.02  & 1    & 9   \\
            Max & 216  & 1.62\,M & 145.6 & 14.89 & 19   & 536 \\
        \bottomrule
    \end{tabular*}
    \caption{Detailed inference costs and proof metrics across the problems resolved by MechMath on number theory subsets in ProverBench.}
    \label{tab:proverbench_full}
\end{table}

\section{Detailed Analysis of Experiments}

To address the concern that localized repair mechanisms (e.g., the Sorrifier) might struggle to maintain global inductive invariants in highly structural paradigms like mathematical induction, MechMath effectively mitigates this potential limitation through rigorous verification of informal proofs and precise evaluation of formal subgoals. 
As shown in Table~\ref{tab:minif2f_categories}, the system achieves consistent and exceptionally high pass rates across various heterogeneous subsets of miniF2F, notably securing an 8/8 success rate on induction problems that require global planning. 
Furthermore, the system performs exceptionally well on the Putnam and IMO 2025 benchmarks, which encompass a diverse range of distinct mathematical disciplines such as continuous real analysis, complex combinatorics, structural discrete geometry, and game theory, thereby demonstrating profound cross-domain robustness.

Besides, In complex proofs, the agent might inadvertently deduce logically flawed subgoals. 
To avoid pursuing unprovable structures, MechMath employs a dual-layer insurance mechanism comprising verification for informal sketches and subgoal assessments, which is further discussed in Appendix~\ref{app:global_flaw}. 
We also provide more detailed analysis of Sorrifier mechanism in Appendix~\ref{app:sorrifier_scope}

\begin{table}[t]
    \centering
    \footnotesize
    \begin{tabular*}{\linewidth}
    {@{\hspace{5pt}\extracolsep{\fill}} l c c c c c c c @{\hspace{5pt}}}
        \toprule
         & IMO & AIME & AMC & Algebra & Number Theory & Induction & Overall\\
         \midrule
        Subset & 19 / 20 & 15 / 15 & 45 / 45 & 88 / 88 & 67 / 68 & 8 / 8 & 242 / 244\\
        \bottomrule
    \end{tabular*}
    \caption{Pass rates across diverse categories on miniF2F}
    \label{tab:minif2f_categories}
\end{table}

\subsection{Empirical Evaluation of Proof Structure Preservation}
\label{app:structural_preservation}

In terms of structural preservation, the Sorrifier maximizes the utility of the already-generated correct proof scaffold during local repairs. Empirically, across 104 successful invocations on miniF2F and 135 on ProverBench, the mechanism achieved an average reuse rate of 74.5\% and 71.1\% of the original proof skeleton lines, respectively, with individual case reuse rates spanning from 5\% to 98\%. This quantitative resilience underscores the efficiency of our decomposition strategy; it establishes a critical algorithmic safeguard that prevents extensive segments of sound mathematical reasoning from being completely discarded due to isolated, low-level tactical failures.

\subsection{Empirical Evaluation of Subgoal Extraction}
\label{app:global_flaw}

To provide a transparent analysis of the failure modes within the Sorrifier mechanism and the extraction pipeline, this section presents a quantitative breakdown of subgoal anomalies across different benchmarks. As described in Section~\ref{main:subgoal_split}, the Reasoner is explicitly prompted to identify which hypotheses and contextual conditions in the current tactic state are relevant to the proof goal, discarding only those that are demonstrably irrelevant. This step is designed precisely to prevent the omission of essential assumptions. More importantly, our pipeline does not treat subgoal extraction as a one-shot process. Each extracted subgoal must undergo a two-stage evaluation by the Verifier before being integrated into the proof pipeline:
\begin{enumerate}[leftmargin=*]
    \item \textbf{Syntactic Validation:} Utilizing Lean checks whether the extracted subgoals are well-formed and adhere to the syntactic conventions of the formal language, ensuring that the subgoals can be processed without compilation errors.
    \item \textbf{Semantic Evaluation:} This stage assesses extraction correctness and logical soundness. Extraction correctness evaluates whether the subgoals accurately reflect the intended reasoning steps, while logical soundness checks whether the subgoal can be proven, typically by examining whether there are obvious counterexamples or essential hypotheses missing.
\end{enumerate}
Subgoals that fail either stage are rejected and returned to the Reasoner for re-extraction. This two-stage evaluation loop serves as a safety net against both missing assumptions and misaligned subtheorems, preventing flawed subgoals from entering the subsequent proof stages and triggering recursive failures or inflated search costs.

The proportion of all three types of errors occurring in all subgoals is summarized in Table~\ref{tab:subgoal_error_breakdown}. The experimental result shows that the error rates of generated subgoals remain minimal on simpler benchmarks like \texttt{miniF2F}, but naturally rise when confronting much harder problems such as \texttt{Putnam 2025} and \texttt{IMO 2025}. Crucially, despite this increase in subgoal anomalies on harder tasks, MechMath consistently maintains its state-of-the-art (SOTA) success rates across the board. This resilient performance demonstrates that the dual-stage evaluation loop effectively intercepts potential logical errors and defects, preventing flawed reasoning from polluting the final proof stream.

\begin{table}[htbp]
    \small
    \centering
    \begin{tabular*}{.99\linewidth}{@{\hspace{5pt}\extracolsep{\fill}} l c c c c @{\hspace{5pt}}}
        \toprule
        Error Type & Putnam 2025 & IMO 2025 & miniF2F & ProverBench \\
        \midrule
        Syntax     & 0.61\% & 5.60\% & 0.41\% & 5.13\% \\
        Extraction & 3.05\% & 4.85\% & 0.00\% & 1.92\% \\
        Logic      & 1.22\% & 1.49\% & 1.23\% & 7.05\% \\
        \bottomrule
    \end{tabular*}
    \caption{Quantitative breakdown of error types occurring across all generated subgoals across different benchmarks.}
    \label{tab:subgoal_error_breakdown}
\end{table}

\subsection{Scope of the Sorrifier and Global Regeneration Boundaries}

\label{app:sorrifier_scope}

To strike an optimal balance between repair efficiency and search exploration, MechMath implements a progressive ``local-decomposition-first, global-regeneration-second'' strategy. Upon encountering a proof failure, the system avoids immediately resorting to expensive global reconstruction; instead, it preferentially invokes the Sorrifier to isolate and salvage the localized failure region. Global regeneration is strictly treated as a tactical fallback, executed only when a defect stems from a foundational structural flaw or when local decomposition triggers a severe logical rejection during verification.

Specifically, we define clear functional boundaries based on the nature of proof anomalies, which directly correspond to different runtime decisions of the system:
\begin{itemize}[leftmargin=*]
    \item \textbf{F1 --- Leaf-step failure:} In this scenario, the global scaffold and intermediate assertions are entirely sound; the failure is restricted to a single tactic or an unclosable low-level code block. The Sorrifier exhibits maximum efficiency here, precisely localizing the exact block, replacing it with a \texttt{sorry} macro, and extracting it as a self-contained independent subgoal to maximize the reuse of the existing scaffold.
    \item \textbf{F2 --- Unsound subgoal / False intermediate claim:} This occurs when an intermediate \texttt{have} assertion is mathematically false or leads to a logical dead end. In this tier, the Sorrifier's role shifts from ``local repair'' to \textbf{``precise diagnosis''}---it generates a critical \texttt{LOGIC\_ERROR} signal to halt downstream error propagation, forcing the pipeline to backtrack and invoke global regeneration equipped with targeted historical hints.
    \item \textbf{F3 --- Structural / Skeleton defect:} This represents the highest tier of strategic missteps, such as selecting an incorrect induction variable, invalid induction hypotheses, or flawed case splits. Because these defects disrupt the core mathematical architecture and sit entirely beyond the scope of the Sorrifier's local processing, addressing F3 flaws relies exclusively on the upstream module executing a thorough global skeleton reconstruction.
\end{itemize}

To demonstrate the practical efficacy of the system when confronting F2-level logical flaws, we first introduce the interception mechanism against erroneous global or intermediate structures. During the subgoal split phase, the subgoal assessment serves as the core line of defense against logical deviations. If the context extracted by the Reasoner is incomplete, the Verifier flags an \texttt{extract error} to prompt re-extraction. More importantly, if the extracted subgoal is diagnosed with a \texttt{logic error} (meaning the premises contradict the goal or cannot entail it), the agent is explicitly prohibited from attempting futile local repairs; instead, the system triggers an \texttt{error fix} state on the overarching formal proof body, forcing it to correct the underlying logical flaw.

This mechanism plays a critical interception role during the empirical evaluation of the challenging \textbf{Putnam 2025 A2} problem. During the proof process, the agent proposes an intermediate bound requiring it to prove the following inequality:

$$\frac{1}{\pi} y (\pi - y) \le \frac{2}{\pi} y , \quad \forall y \in \left[0, \frac{\pi}{2}\right]$$

The Verifier mathematically invalidates this inequality through counterexample computation (e.g., substituting $y=0.5$ yields a left-hand side of $0.421$, which is greater than the right-hand side of $0.318$) and decisively flagged it as a \texttt{logic error}. This successfully intercepts a doomed local proof search and forced the pipeline to backtrack to the formal proof spine for structural reorganization.

Empirical statistics on the miniF2F benchmark further confirm the necessity of this module: across the evaluation, the Verifier successfully intercepts a \texttt{logic error} 5 times across 3 distinct problems and corrected an \texttt{extract error} once. By seamlessly integrating rigorous upstream global planning with real-time logical scrutiny of formal subgoals, MechMath effectively neutralizes theoretical and logical defects.

Furthermore, to clearly delineate the dynamic flow between local repair (Sorrifier) and global regeneration when confronting intertwined high-level F3 skeleton defects and F1 leaf-step failures, we provide an induction-heavy case study (from ProverBench, \texttt{number\_theory\_\_p31}) that relies heavily on global structure.

\paragraph{Stage 1 --- Wrong induction hypothesis (F3-level global skeleton defect)}
The agent initially attempted strong induction directly on the predicate. Because the subsequent steps require applying the induction hypothesis to \texttt{n.minFac}, and the current induction hypothesis (\texttt{ih}) fails to expose the statement in this form, the inductive step cannot be closed. This represents an architecture-breaking F3-level defect where no leaf-level \texttt{sorry} replacement can recover from the directional misstep.

\begin{lstlisting}[frame = single, style=lean]
  induction n using Nat.strongInductionOn with
  | ind n ih =>
    intro hn
    let a := n.minFac
    have ha_prime : a.Prime := Nat.minFac_prime (by omega)
    have ha_dvd : a ∣ n := Nat.minFac_dvd n   -- Need IH on n/a, but ih cannot provide it
\end{lstlisting}

\paragraph{Stage 2 --- Regenerated correct skeleton}
Since the proof cannot be locally patched, the system decisively abandons localized repair and triggers global regeneration. By restructuring the proof architecture and strengthening the induction hypothesis, the system introduces an explicit bound $n$ and inducts on it, successfully generalizing the claim to hold for all $m \le n$. At this point, the global induction skeleton becomes completely sound.

\begin{lstlisting}[frame = single, style=lean]
  have H3_aux : ∀ n m : ℕ, m ≤ n → m % 4 = 3 → ∃ q : ℕ, q.Prime ∧ q % 4 = 3 ∧ q ∣ m := by
    intro n
    induction n with
    | zero =>
      intro m hm_le hm_mod
      have : m = 0 := by omega   -- Base case: vacuous
      omega
\end{lstlisting}

\paragraph{Stage 3 --- Sorrifier local repair (F1-level leaf-step failure)}
Armed with a correct global induction skeleton, the remaining failure reduces to a single isolated leaf fact, downgrading the error to an F1 tier. Here, the Sorrifier perfectly preserves the fully generated induction architecture and replaces \textbf{only} the failing leaf step with a \texttt{sorry} macro, extracting it as a standalone subgoal lemma to be resolved independently, thereby eliminating the computational overhead of redundant global reconstruction.

\begin{lstlisting}[frame = single, style=lean]
  have hm_cases : m = n + 1 ∨ m ≤ n := by omega
  rcases hm_cases with rfl | m_le_n
  · let a := (n + 1).minFac
    have ha_prime : a.Prime := Nat.minFac_prime (by omega)
    obtain ⟨b, hab⟩ : ∃ b, n + 1 = a * b := Nat.minFac_dvd (n + 1)
    have hb_pos : 0 < b := by sorry   -- <- Localized leaf gap; the scaffold above is perfectly reused
\end{lstlisting}

\end{document}